\documentclass[10pt,twocolumn]{IEEEtran}
\usepackage{mathrsfs}
\usepackage{amssymb}
\usepackage{amsmath}
\usepackage{bm}
\usepackage{epsfig}
\usepackage{color}
\usepackage{csquotes}
\usepackage[font=small]{caption}
\usepackage{subcaption}
\usepackage{gensymb}
\usepackage{float}
\usepackage{cite}
\usepackage{xcolor}

\newtheorem{theorem}{Theorem}

\newtheorem{lemma}{Lemma}
\newtheorem{remark}{Remark}
\newtheorem{proposition}{Proposition}

\newtheorem{assumption}{Assumption}

\DeclareMathOperator*{\argmax}{arg\,max}
\usepackage{accents}
\newcommand{\ubar}[1]{\underaccent{\bar}{#1}}

\usepackage[labelformat=simple]{subcaption}

\usepackage[ruled,vlined]{algorithm2e}
\usepackage[normalem]{ulem}
\SetArgSty{textnormal}
\SetKwInput{KwResult}{Initialization}

\begin{document}

\title{Multi-Agent On-Line Extremum Seeking Using Bandit Algorithm}
\author{ Bin~Du$^{\dagger}$,
Kun~Qian$^{\ddagger}$,
Christian~Claudel$^{\ddagger}$,
and~Dengfeng Sun$^{\dagger}$
\thanks{$^{\dagger}$Bin Du is Ph.D. student, and Dengfeng Sun is Professor, with the School of Aeronautics and Astronautics, Purdue University, West~Lafayette,~IN 47907, { \{du185, dsun\}@purdue.edu}}%
\thanks{$^{\ddagger}$Kun Qian is Ph.D. student, and Christian Claudel is Associate Professor, with the Department  of Civil, Architectural, and Environmental Engineering, the University of Texas at Austin, Austin, TX 78712, { \{kunqian, christian. claudel\}@utexas.edu }}%
\thanks{Bin Du and Kun Qian contributed equally to this manuscript.}
}
\maketitle

\begin{abstract}
This paper presents a learning based distributed algorithm for solving the on-line extremum seeking problem with a multi-agent system in an unknown dynamical environment. Our algorithm, building on a novel notion termed as \textit{dummy~confidence~upper bound} (D-UCB), integrates both estimation of the unknown environment and task planning for the multiple agents simultaneously, and as a consequence, enables the multi-agent system to track the extremum spots of the dynamical environment in an on-line manner. Unlike the standard confidence upper bound (UCB) algorithm in the context of multi-armed bandits, the introduction of D-UCB significantly reduces the computational complexity in solving subproblems of the multi-agent task planning, and thus renders our algorithm exceptionally computation-efficient in the distributed setting. The performance of the algorithm is theoretically guaranteed by showing a sub-linear upper bound of the cumulative regret. Numerical results on a real-world pollution monitoring and tracking problem are also provided to demonstrate the effectiveness of our algorithm.
\end{abstract}

\section{Introduction}
Over the last few decades, extremum seeking, also known as source seeking, has~been a fundamentally crucial problem and attracted increasing~attention, due to its numerous applications including surveillance\mbox{\cite{tang2005motion, ghaffarkhah2012path}}, environment and health \mbox{monitoring \cite{lu2012spoc,lu2016cooperative,qian2020real, mascarich2018radiation}}, disaster response~\cite{sugiyama2013real,arnold2018search}, to name a~few. Extremum seeking involves locating one or several spots, associated with the maximum/minimum values of interest, in a possibly unknown and noisy environment. Oftentimes, those extremum spots are of particular importance in many real-world applications. For instance, in the scenario of flood/tide monitoring\cite{qian2020real,abdelkader2013uav}, paying specific attention to the extremum spots, which usually correspond to the flood peaks, could provide stake holders with timely warnings. In this paper, we are particularly interested in solving the problem of extremum seeking with a multi-agent system, in which a network of agents are deployed and expected to cooperatively locate as many extremum spots as possible. It is highlighted that the underlying environment considered in this paper is not only unknown but also dynamically changing as the multiple agents acquire knowledge from it. Under such a circumstance, the agents need to collaboratively explore the unknown environment and simultaneously track the dynamically changing extremum spots. We remark that these two settings, i.e., the multi-agent system and dynamical environment, make our problem significantly challenging to solve.

Indeed, there have been various existing works~{\cite{mascarich2018radiation,rolf2020successive,li2014cooperative,fabbiano2014source,brinon2015distributed,fabbiano2016distributed,atanasov2012stochastic, azuma2012stochastic,atanasov2015distributed}} studying the extremum seeking problem in both centralized and distributed settings. The predominate approaches to this problem are typically based on the gradient estimation, i.e., driving the agent(s) to trace along with the estimated gradient direction toward the target which is usually associated with local extremum values. {In particular, the authors in~\cite{li2014cooperative} designed the distributed source seeking control law for a~group of cooperative robots by modeling the unknown environment as a time-invariant and concave real-valued function.~Besides, the diffusion process is considered in~\cite{fabbiano2014source} for the scenarios of dynamically environment. The authors in~\cite{brinon2015distributed,fabbiano2016distributed} also studied the distributed source seeking problem by forcing the multiple agents to follow a circular formation. In addition, the stochastic gradient based methods are further proposed in~\cite{atanasov2012stochastic, azuma2012stochastic,atanasov2015distributed} to drive the single robot or robot network to the desired targets.} All these gradient based extremum seeking methods are closely related to the first-order optimization algorithm, and their advantages are often attributed to the fact that only local measurements are required during the whole seeking process without the need of knowing the agent's global positional information (GPS is thus denied). Nevertheless, we should note that, also inherited from the first-order optimization algorithm, these gradient based methods are very likely to stuck at the local extremum points when the considered environment is non-convex/non-concave. More importantly, the estimation of gradients is usually sensitive to the noise presented in measurement and/or the underlying environment, and thus some other assumptions regarding the noise need to be imposed in the problem setup.

In order to address the aforementioned issues, a very recent approach, which is closely related to our ideas, devises a learning based adaptive scheme in~\cite{rolf2020successive}, by leveraging the notion of UCB in the study of multi-armed bandits algorithms. This approach, termed as \texttt{AdaSearch}, maintains a set of candidate points which are likely to be the extremum spots, and let the agent repeat a predetermined trajectory so that it can adaptively collect information from the unknown environment and iteratively update the candidate set. As a consequence, the agent will be able to eventually identify the desired extremum spots after sufficient information is acquired. However, we should remark that there are two potential drawbacks of the \texttt{AdaSearch} scheme: 1) it requires the agent to strictly follow the predetermined trajectory, which might be inefficient at the later stage of the algorithm; and 2) only one single agent is considered and the static environment is presumed, thus it is not applicable in our problem setup while considering the multi-agent system and dynamical environment.

\begin{figure}
  \centering
  \includegraphics[width=0.95\linewidth]{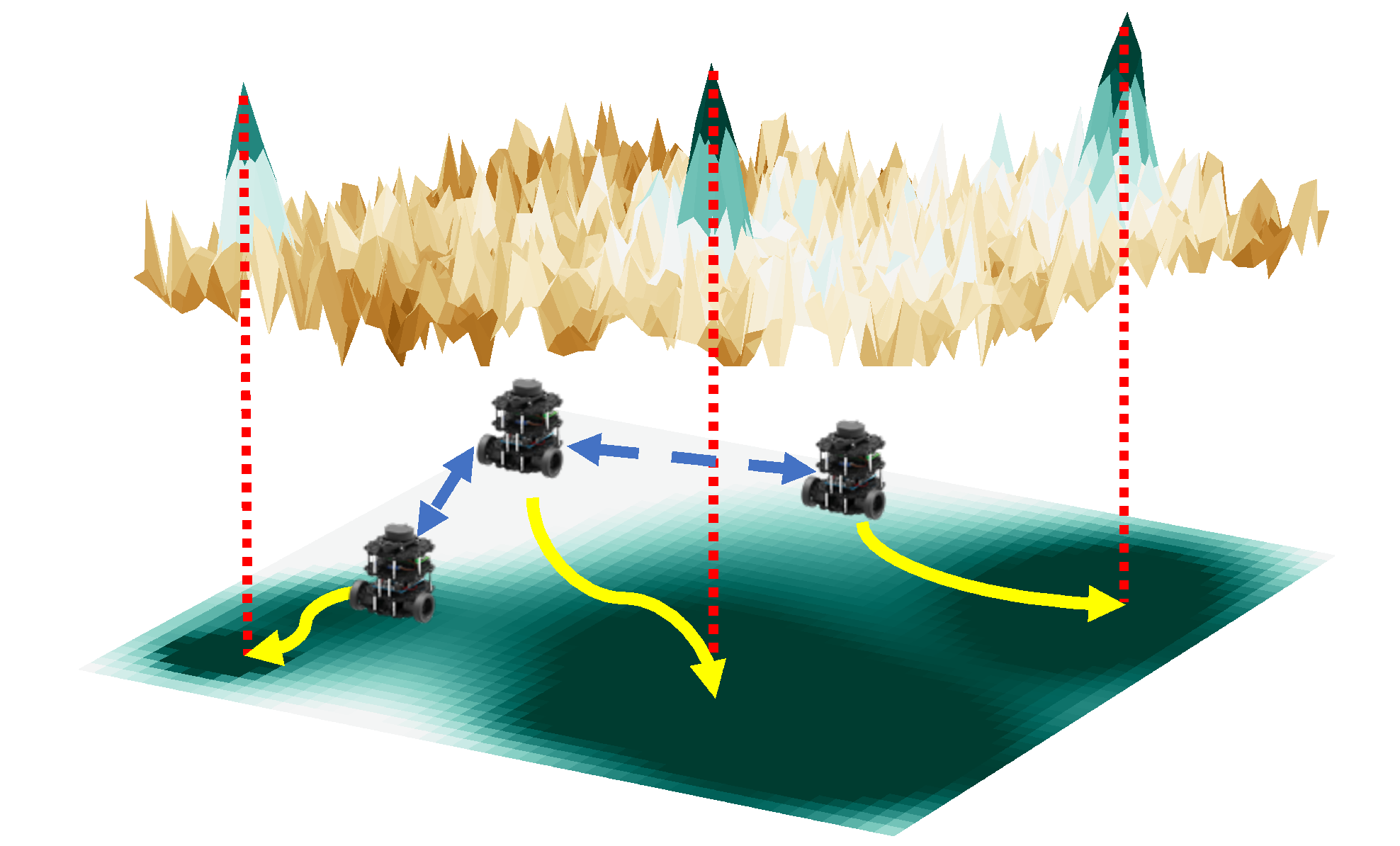}
  \caption{\small Visualization of the proposed extremum seeking approach: the lower layer corresponds to the unknown environment that needs to be explored; the upper layer depicts the D-UCB which guides the agents' task planning. Each agent exchanges information with its immediate neighbors and cooperatively estimate the unknown environment.}
  \vspace{-10pt}
  \label{fig:demo}
\end{figure}

Inspired by~\cite{rolf2020successive}, in this paper we also develop a learning based algorithm by integrating the estimation of unknown environment and task planning for multi-agent simultaneously. Nevertheless, in contrast to the \texttt{AdaSearch} scheme, we here let the agents cooperatively determine their paths by themselves, and introduce the novel variant of UCB, namely D-UCB, which greatly helps reduce the computational complexity in solving multi-agent task planning problems. These two points also make our algorithm implementable in both distributed and on-line manners. In addition, other differences between this paper and~\cite{rolf2020successive} are also noteworthy:~1) while the measurement noise is assumed to follow a Poisson process in~\cite{rolf2020successive}, we consider the noise to be Gaussian distributed; see~Sec.~\ref{subsec:sourceSeekingwithEstimates}; and 2) the \texttt{AdaSearch} scheme utilizes both lower and upper confidence bounds to guide the agent's decision, in contrast, we only need to compute the upper bound with our algorithm. The mechanism of our algorithm is illustrated in Fig.~\ref{fig:demo}.

It is worth noting that the idea of UCB has been commonly adopted in solving the relevant problems, such as environment monitoring~\cite{marchant2012bayesian, bai2016information, miller2015ergodic}, sensor coverage~\cite{luo2018adaptive, luo2019distributed, benevento2020multi} and so on. In these problems, the environment is often modeled as a Gaussian process~\cite{o1978curve}. However, as suggested in~\cite{rolf2020successive} and also in~\cite{o1978curve} itself, such a modeling strategy often imposes to some extent the assumption of smoothness of the underlying environment. Therefore, it may not be able to reflect some specific scenarios of the extremum seeking problem; for example, when considering the sparse, heterogeneous emission encountered in the radiation detection. On this basis, in this paper we apply a generic state-space model for the dynamical environment; see details in {Sec.~\ref{subsec:sourceSeekingwithEstimates}}. {Furthermore, when it comes to the distributed setting, solving the standard UCB based maximization is essentially of combinatorial nature and thus can be extremely complicated to find the exact solutions. In order to cope with such an issue, our idea of D-UCB helps decompose the maximization problems marginally.} This also makes our work significantly different with other literature relying on the standard UCB approach. 

{The rest of this paper is organized as follows. Section II formally defines the considered distributed extremum seeking problem involved with the estimation of the unknown environment. Section III develops our distributed on-line algorithm and Section IV presents the simulation results to demonstrate effectiveness of the algorithm. Lastly, Section V concludes this paper. For the reader’s convenience, the proofs of proposition and theorem are provided in Appendix. We should note that an earlier version of this paper appears in~\cite{2021ICRA}, but the present paper has been significantly enhanced, including the detailed theoretical proofs, more comprehensive interpretation of the proposed algorithm, and more extensive numerical results by considering a real-world pollution monitoring and tracking application.}

\section{Problem Statement}

\subsection{Distributed Extremum Seeking}

In this subsection, we formalize the problem of distributed extremum seeking with the multi-agent system. 
Let us consider a bounded and {obstacle-free} environment, in which the extremum spots of interest are present. In particular, we specify the considered environment by a set of points $\mathcal{S}$ with each element $\mathbf{s} \in \mathcal{S}$ representing the position of the point. Since the environment has been assumed to be bounded, it is easy to see that the set $\mathcal{S}$ is finite. We denote $N$ the number of points in the set, i.e., $N = |\mathcal{S}|$. For each point~$\mathbf{s}$ in $\mathcal{S}$, there exists a real-valued function $\phi_k(\cdot): \mathcal{S} \to \mathbb{R}_+$ that maps the point's positional information $\mathbf{s}$ to a positive quantity $\phi_k(\mathbf{s})$ indicating the value of field at the time-step $k$. Naturally, in order to locate the extremum spots, our objective is to deploy the multiple agents to the points with the highest quantities $\phi_k(\mathbf{s})$. More precisely, we employ a network of~$I$ agents which are capable of moving among $\mathcal{S}$ and communicating with other connected neighbors, and expect them to track as many extremum spots as possible. That is, at each time-step $k$, each individual agent $i \in \mathcal{I}:=\{1,2,\cdots,I\}$ aims at seeking its best position $\mathbf{p}_k^\star[i] \in \mathcal{S}$ by cooperatively solving the following maximization problem,


\begin{equation}\label{basicDCM}
\begin{aligned}
\mathop {\text{maximize} }\limits_{\mathbf{p}[i] \in \mathcal{S},\, i \in \mathcal{I}}  \quad &F_k(\mathbf{p}[1],\mathbf{p}[2],\cdots, \mathbf{p}[I]) = \sum_{\mathbf{s} \in \cup_{i=1}^I\mathbf{p}[i]} \phi_k(\mathbf{s}).
\end{aligned}
\end{equation}
Note that the objective function $F_k(\cdot) : \mathcal{S}^I \to \mathbb{R}_+$ maps the agents' positions $\mathbf{p}[i]$'s to a positive scalar that sums all \textit{distinct} measured quantities. Throughout this paper, we assume that the maximizer $\big(\mathbf{p}_k^\star[1],\mathbf{p}_k^\star[2],\cdots, \mathbf{p}_k^\star[I]\big)$ of problem~\eqref{basicDCM} is unique at each time-step $k$ and express it as a compact form $\mathbf{p}_k^\star = \big[\mathbf{p}_k^\star[1],\mathbf{p}_k^\star[2],\cdots, \mathbf{p}_k^\star[I]\big] \in \mathcal{S}^I$.

It should be noted that, since the set $\mathcal{S}$ is finite, the above maximization problem can be naively solved by assigning the \mbox{$i$-th} agent to the point $\mathbf{p}[i]$ which has the $i$-th largest quantity $\phi_k\big(\mathbf{p}[i]\big)$. However, such a naive scheme inherently assumes each agent to be aware of its exclusive global ID which is a restrictive requirement in a fully distributed architecture~\cite{ould2009distributed}. As an alternative way to solve the optimization problem~\eqref{basicDCM}, we shall remark that the problem can be viewed as a special case of the monotone submodular maximization, and thus can be solved by the distributed algorithm proposed in our previous work~\cite{du2020jacobi}. The key idea of this algorithm is to find the equilibrium solution, and interestingly, it can be verified that the problem~\eqref{basicDCM} has a unique equilibrium which is coincident with the optimal solution. We refer the interested reader to our work~\cite{du2020jacobi} for details on the distributed~algorithm.

\subsection{Extremum Seeking via Estimation on the Environment}\label{subsec:sourceSeekingwithEstimates}

Notice that the problem~\eqref{basicDCM} considered in the previous subsection is somewhat trivial, since it implicitly assumes that each agent perfectly knows the state $\phi_k(\mathbf{s})$ of the entire environment at each time-step $k$. This is unrealistic for the real-world applications. On this account, we next let the network of agents cooperatively estimate the environment based on the local noisy measurements, and in the following, we first introduce the dynamics of the environment states as well as the measurement model of the agents.

Suppose that the vector $\bm{\phi}_k \in \mathbb{R}_+^N$ stacks each individual state $\phi_k(\mathbf{s})$ for all points $\mathbf{s}$ in the environment $\mathcal{S}$. We consider the following 
linear time-varying (LTV) model for the environment state, i.e.,
\begin{align}\label{linearModel}
  {\bm\phi_{k+1} = A_{k+1} \bm\phi_{k}},
\end{align}
where $A_k \in \mathbb{R}^{N \times N}$ denotes the state transition matrix. In order to ensure that the above maximization problem~\eqref{basicDCM} is well-defined, it is required to guarantee that the state $\bm\phi_k$ is always bounded and also will not vanish to zero as the time-step $k$ increases. More precisely, we use the following assumption to constrain the behavior of the state dynamics.

\begin{assumption}\label{assump:dynamicsBound}
  For the LTV model~\eqref{linearModel}, there exist uniform lower and upper bounds $0 < \ubar{\alpha}\le \bar{\alpha} < \infty $ such that, for $\forall k \ge t > 0$,
  \begin{align}
    \ubar{\alpha} \cdot \mathbf{I} \le A[k:t]^\top A[k:t] \le \bar{\alpha} \cdot \mathbf{I},
  \end{align}
  where $\mathbf{I}$ denotes the identity matrix with appropriate dimensions and the state propagation matrix $A[k:t] \in \mathbb{R}^{N \times N}$ is written as
  \begin{align}
    A[k:t] = A_k A_{k-1} \cdots A_t.
  \end{align}
\end{assumption}

\begin{remark}
  Note that the above Assumption~\ref{assump:dynamicsBound} is reasonably required to ensure that the maximum components of $\bm{\phi}_k$ are always recognizable for the multiple agents. Moreover, this assumption also implies the invertibility of the matrices $A_k$'s. In fact, as suggested in~\cite{li2019boundedness} (see Remark~2), for the sampled-data system (one of the mostly studied discrete-time systems), the matrix $A_k$ is naturally invertible since it is often obtained by discretization of the continuous-time system. Such an assumption has been quite standard in various research studying the state estimation problems, see e.g., \cite{li2019boundedness,battistelli2014kullback,battistelli2014consensus,cattivelli2010diffusion}.
\end{remark}

In addition, we consider the following linear stochastic measurement model for each agent~$i$,
\begin{align}\label{dynMeasurement}
  \mathbf{z}^i_k = H^i\big(\mathbf{p}_k[i]\big) \bm{\phi}_k + \mathbf{n}_k^i.
\end{align}
where $\mathbf{z}_k^i\in \mathbb{R}^{m}$ represents the measurement~obtained by the agent $i$ at the time-step $k$\footnote{For simplicity, we assume that each sensor's measurement has the same dimension $m$; this can be easily relaxed to a general case.}; $H^i\big(\mathbf{p}_k[i]\big) \in\mathbb{R}^{m\times N}$ denotes the measurement matrix depending on the agent's position $\mathbf{p}_k[i]$; and $\mathbf{n}_k^i \in \mathbb{R}^{m}$ is~corresponding to the measurement noise satisfying the following assumption.
\begin{assumption}\label{assump:measurementNoise}
  It is assumed that the  measurement noise $\mathbf{n}_k^i$ follows the independent and identically distributed (\textit{i.i.d.}) Gaussian for each individual agent $i$, with zero-mean and~covariance matrix $V^i = v^i\cdot \mathbf{I}$. In addition, there exist lower and upper bounds $0 < \ubar{v}\le \bar{v} < \infty $ such that
  \begin{align}
    \ubar{v} \le v^i \le \bar{v}, \, \forall i \in \mathcal{I}.
  \end{align}
\end{assumption}

\begin{remark}\label{remark:measurement}
  We shall remark that the measurement matrix $H^i\big(\mathbf{p}_k[i]\big)$ is not specified in the above model~\eqref{dynMeasurement}. In fact, it can be defined by various means based on the agent's position. One of the simplest way is to let $H^i\big(\mathbf{p}_k[i]\big) = \mathbf{e}_l^\top$ where $\mathbf{e}_l \in \mathbb{R}^N$ is an unit vector, i.e., the $l$-th column of the identity matrix, and $l\in \{1, 2,\cdots, N\}$ denotes the index of the position $\mathbf{p}_k[i]$ in the environment $\mathcal{S}$. This means that the agent only measures the quantity at the point where it currently locates. Such a choice of $H^i\big(\mathbf{p}_k[i]\big)$ is actually adopted in~\cite{rolf2020successive} as the so-called point-wise sensing model. Besides, some other specifications of the measurement matrix are also used in the existing works. For instance, a circular sensing area with radius~$r_i$ is applied in \cite{habibi2016gradient}, which implies that,
  \begin{align}\label{circularMeasure}
    H^i\big(\mathbf{p}_k[i]\big) = \big[\mathbf{e}_l\big]^\top_{l \in \mathcal{C}_k^i},
   \end{align} 
   where the set $\mathcal{C}^i_k:= \{l\;|\; \|{\mathbf{s}_l - \mathbf{p}_k[i]}\| \le r^i\}$ includes the indices of all points $\mathbf{s}_l$ that fall into the disk which is centered at $\mathbf{p}_k[i]$ and has radius $r^i$.
\end{remark}

Based on the measurement model~\eqref{dynMeasurement}, one should notice that, when some mild conditions on the measurement matrices are satisfied, the true value of $\bm{\phi}_k$ can be estimated by many techniques, such as least-squares, Kalman filter, to name a few. {Therefore, the problem of distributed extremum seeking with an unknown environment can be addressed by a simple approach which contains the following two phases separately: 1) let the network of agents move around the environment and obtain an accurate enough estimation of the state; and 2) specify the agents' target positions at each time-step $k$ by solving the maximization problem~\eqref{basicDCM} based on the estimated states. However, this is essentially an off-line approach, since the agents do not have specific targets when estimating the environment in the phase 1) and the phase 2) cannot be started until an accurate enough estimate is obtained. Motivated by this, in the next section, we aim to integrate the above two phases together and propose an adaptive on-line framework. That is, the agents recursively update their target positions; meanwhile, measure and estimate the unknown environment, until the objective is reached in which the network of $I$ agents manages to track the moving extremum spots.} 

\section{An Adaptive On-line Framework}\label{sec:onlineAlgo}
 

\subsection{Kalman Consensus Filter}\label{subsec:kalman}

Let us begin by rewriting the measurement model~\eqref{dynMeasurement} into the following compact form
\begin{align}
  \mathbf{z}_k = H_k\bm{\phi}_k + \mathbf{n}_k.
\end{align}
Note that here \mbox{$\mathbf{z}_k = [(\mathbf{z}^1_k)^\top, (\mathbf{z}^2_k)^\top, \cdots, (\mathbf{z}^I_k)^\top]^\top \in \mathbb{R}^M$} is the measurement obtained by all agents with dimension \mbox{$M = mI$}; \mbox{$H_k =[H^1(\mathbf{p}_k[1])^\top\hspace{-2pt}, H^2(\mathbf{p}_k[2])^\top, \hspace{-2pt}\cdots\hspace{-2pt}, H^I(\mathbf{p}_k[I])^\top]^\top \hspace{-2pt}\in\hspace{-2pt} \mathbb{R}^{M \times N}$} stacks all local measurement matrices as a collective global one\footnote{When writing $H_k$, with slight abuse of notation, we have absorbed the dependency on the agents' positions $\mathbf{p}_k[i]$'s into the index $k$.}; and $\mathbf{n}_k = [(\mathbf{n}^1_k)^\top, (\mathbf{n}^2_k)^\top, \cdots, (\mathbf{n}^I_k)^\top]^\top \in \mathbb{R}^M$ denotes the Gaussian noise with zero-mean and covariance matrix expressed as 
\begin{align}\label{covMat}
  V := \text{Diag}\{V^1, V^2, \cdots,V^I\} \in \mathbb{R}^{M \times M}.
\end{align}
Subsequently, the centralized Kalman filter for estimating the mean $\widehat{\bm{\phi}}_k \in \mathbb{R}^N$ and covariance $\Sigma_k \in\mathbb{R}^{N\times N}$ performs the following recursions,
\begin{subequations}\label{Kalman}
  \begin{align}
    \Sigma_{k+1} &= A_{k+1}\big(\Sigma_{k}^{-1} + Y_k \big)^{-1}A_{k+1}^\top;\label{KalmanSigma}\\
    \widehat{\bm{\phi}}_{k+1} &= A_{k+1}\Big(\widehat{\bm{\phi}}_{k} + (\Sigma_{k}^{-1}+Y_k)^{-1}(\mathbf{y}_k - Y_k\widehat{\bm{\phi}}_{k})\Big)\label{KalmanPhi},
  \end{align}
 \end{subequations} 
where the two variables $Y_k := (H_k)^\top V^{-1}H_k \in\mathbb{R}^{N\times N}$ and $\mathbf{y}_k:=(H_k)^\top V^{-1}\mathbf{z}_k \in \mathbb{R}^N$, often referred to as the new information, incorporate the measurements into the updates. 

It is worth mentioning that the Kalman filter~\eqref{Kalman} readily estimates the unknown environment in the desired on-line manner, i.e., the multiple agents move to new positions, obtain the new measurements, and  then update their estimates of the environment. {However, we should note that two issues may arise: i) the statistical property of the classical Kalman filter may no longer hold due to the sequential decision process}; ii) such an on-line procedure is performed in a centralized way, since the new information $Y_k$ and $\mathbf{y}_k$ are involved with the data obtained/maintained by all agents. In order to devise a distributed scheme to run the Kalman filter~\eqref{Kalman}, many existing works, \mbox{e.g.,~\cite{olfati2005distributed, olfati2005consensus, olfati2007distributed}}, leverage the special structure of the noise covariance $V$. Considering the diagonal structure of the matrix $V$, as shown in~\eqref{covMat}, the new information can be further expressed as
\begin{subequations}\label{newInfo}
 	\begin{align}
 		Y_k &= \sum_{i=1}^I H^i(\mathbf{p}_k[i])(V^i)^{-1}H^i(\mathbf{p}_k[i])^\top;\\
 		\mathbf{y}_k &= \sum_{i=1}^I H^i(\mathbf{p}_k[i])(V^i)^{-1}\mathbf{z}^i_k,
 	\end{align}
 \end{subequations} 
which means that $Y_k$ and $\mathbf{y}_k$ can be computed by simply summing all the local information together. This motivates the development of Kalman consensus filter, in which each agent first carries out an average/sum consensus procedure to fuse local information and then performs the standard Kalman update~\eqref{Kalman}. 

\subsection{The Distributed On-Line Extremum Seeking Algorithm}\label{subsec:UCB}
In the previous subsections, we focused on the estimation of the unknown environment. Our question now becomes: how to integrate the estimation together with the agents' decision-making processes. A naive idea would be using the estimated mean value $\widehat{\bm{\phi}}_k$ at each time-step $k$, and then solving the following maximization problem,
\begin{align}\label{naiveAlgo}
  \mathbf{p}_{k} \in \argmax_{\mathbf{p}[i] \in \mathcal{S},\, i \in \mathcal{I}} \sum_{\mathbf{s} \in \cup_{i=1}^I\mathbf{p}[i]} \widehat{\phi}_k(\mathbf{s}).
\end{align}
 {Here, we use $\widehat{\phi}_k(\mathbf{s}) \in \mathbb{R}$ to denote one component of the vector~$\widehat{\bm{\phi}}_k$ which corresponds to the point $\mathbf{s}$ in the environment. It should be emphasized that such a scheme cannot guarantee the network of agents to track the extremum spots with the highest true $\phi_k(\mathbf{s})$'s. To elaborate on this, let us consider a special case where the environment is static, i.e., $\bm{\phi}_k = \bm{\phi}_0,\forall k\in\mathbb{N}_+$. Subsequently, an undesired but possible scenario is that the agents significantly underestimate the maximum value $\phi_0(\mathbf{s^\star})$ at the initial stage, i.e., $\widehat{\phi}(\mathbf{s^\star}) \ll \phi_0(\mathbf{s^\star})$, and as a result, the agents will never have another chance to visit the key point~$\mathbf{s}^\star$. On this account, it can been seen that merely utilizing the estimated mean is insufficient to drive the network of agents to the desired positions. To address this, we next take advantage of both the estimated mean $\widehat{\bm{\phi}}_k$ and covariance $\Sigma_k$ to develop our {distributed on-line extremum seeking algorithm.}

 Based on $\widehat{\bm{\phi}}_k$ and $\Sigma_k$, let us introduce an additional variable $\bm{\mu}_k \in \mathbb{R}^N$, which we refer to as D-UCB,
 \begin{align}\label{UCB}
  \bm{\mu}_k: = \widehat{\bm{\phi}}_k + \beta_k(\delta) \cdot\text{diag}^{1/2}(\Sigma_k).
 \end{align}
 Note that the operator $\text{diag}^{1/2}(\cdot): \mathbb{R}^{N\times N} \to \mathbb{R}^{N}$ maps the square root of the matrix diagonal elements to a vector, {and the parameter $\beta_k(\delta) >0$ depending on the critical confidence level $\delta$ will be specified later on. In fact, the intuition behind this notion of \mbox{D-UCB} is straightforward: each $\bm{\mu}_k$ provides a probabilistic upper bound of the true value $\bm{\phi}_k$ by utilizing the current mean and covariance. 
 Next, we formalize, with the following proposition, how the true value $\bm{\phi}_k$ is upper bounded by the D-UCB $\bm{\mu}_k$ with the probability related to $\delta$.}

 \begin{proposition}\label{lemma:DUCB}
  Under Assumptions \ref{assump:dynamicsBound} and \ref{assump:measurementNoise}, let the state estimates $\widehat{\bm{\phi}}_k$ and~$\Sigma_k$ be generated by the Kalman (consensus) filter \eqref{Kalman} with the {initialization $\widehat{\bm\phi}_0$ and $\ubar{\sigma}\cdot\mathbf{I}\le\Sigma_0\le\bar{\sigma}\cdot\mathbf{I}$}, then it holds that, for $\forall k >0$,
  \begin{align}\label{IneqUCB}
    {\mathbb{P}\Big(\big|\widehat{\bm{\phi}}_k - \bm{\phi}_k\big| \preceq \beta_k(\delta)\cdot\text{diag}^{1/2}(\Sigma_k) \Big) \ge 1-\delta},
  \end{align}
  where the operators $|\cdot|$ and $\preceq$ are defined element-wise, the probability $\mathbb{P}(\cdot)$ is taken on random noises $(\mathbf{n}_1, \mathbf{n}_2, \cdots,\mathbf{n}_k)$, and $\{\beta_k(\delta)\}_{k\in\mathbb{N}_+}$ is an increasing sequence, defined as
  \begin{align}\label{betaK}
    \beta_k(\delta) \ge N^{3/2}C_1 + N^2 C_2\cdot\sqrt{\log\Big(\frac{\bar{\sigma}/\ubar{\sigma} + \bar{\alpha}\bar{\sigma}\cdot k / \ubar{v}^2}{\delta^{2/N}}\Big)},
  \end{align}
  with $C_1 = \|\widehat{\bm{\phi}}_0 - \bm{\phi}_0\| /\sqrt{\ubar{\sigma}}$ and $C_2 = \bar{v}^2\sqrt{\max\{2,2/\ubar{v}\}}$.
 \end{proposition}

\vspace{10pt}
 \begin{IEEEproof}
 See Appendix~\ref{subsec:appendA}.
 \end{IEEEproof}
 \vspace{10pt}

The above Proposition~\ref{lemma:DUCB} inherently constructs a polytope centered at the state estimate $\widehat{\bm{\phi}}_k$ such that the true state $\bm{\phi}_k$ falls into it with probability at least $1-\delta$. Based on the polytope, it can be seen that the D-UCB~$\bm\mu_k$ takes the upper bounds marginally and each element $\mu_k(\mathbf{s})$ is guaranteed to be satisfied with $\mu_k(\mathbf{s}) \ge \phi_k(\mathbf{s})$ with probability at least~$1-\delta$. 
Consequently, we can use the defined D-UCB $\bm\mu_k$ to update the agents' target positions in the on-line manner, by solving the following maximization problem:
\begin{align}\label{algo}
  \mathbf{p}_{k} \in \argmax_{\mathbf{p}[i] \in \mathcal{S},\, i \in \mathcal{I}} \sum_{\mathbf{s} \in \cup_{i=1}^I\mathbf{p}[i]} {\mu}_k(\mathbf{s}).
\end{align}

It is worth emphasizing that the introduction of D-UCB here helps reduce the computational complexity of the proposed algorithm significantly, when solving the problem in the distributed manner. Since the standard UCB is defined in a joint sense, when solving the multi-agent maximization problem~\eqref{algo} with the standard UCB, it is inherently of combinatorial nature and thus can be extremely complicated to find the exact solution. In contrast, due to the fact that the D-UCB takes the upper bounds marginally here, the maximization~\eqref{algo} can be essentially decomposed and becomes much easier to solve for exact solutions. We remark this as one of the most important contributions of our algorithm.
At last, we summarize our distributed on-line extremum seeking scheme in the following Algorithm~\ref{algo:UCB} and establish its regret analysis as the following theorem.

  \begin{algorithm}
 \SetAlgoLined
 \caption{Distributed On-Line Extremum Seeking}
\vspace{5pt}
  \KwResult{\parbox{\dimexpr\textwidth+2\algomargin\relax}{Each agent $i$ initializes its own estimates }\\
  \parbox{\dimexpr\textwidth-2\algomargin\relax}{$\widehat{\bm\phi}_0$ and $\Sigma_0$, and computes the target position $\mathbf{p}_0[i]$. Set}\\
    \parbox{\dimexpr\textwidth-2\algomargin\relax}{the confidence level $\delta$. Let $k =0$. }\\
  } 
  \vspace{5pt}
  \While{the stopping criteria is NOT satisfied}{
  \vspace{5pt}
  Each agent $i$ \textbf{simultaneously} performs\\
  \vspace{5pt}
  {\texttt{\textbf{Step 1}}} (\textit{Measuring}): Obtain the measurement $\mathbf{z}_i^k$ based on the measurement matrix $H^i(\mathbf{p}_k[i])$;\\
  \vspace{5pt}

   {\texttt{\textbf{Step 2}}} (\textit{Kalman Filtering}): Collect information from neighbors, \hspace{-2pt}obtain mean $\widehat{\bm{\phi}}_{k+1}$ and covariance $\Sigma_{k+1}$ by Kalman consensus filter~\eqref{Kalman};
   \vspace{5pt}

   {\texttt{\textbf{Step 3}}} (\textit{D-UCB Computing}): Compute via~\eqref{UCB} the updated D-UCB $\bm{\mu}_{k+1}$ by $\widehat{\bm{\phi}}_{k+1}$ and $\Sigma_{k+1}$;
   \vspace{5pt}

   {\texttt{\textbf{Step 4}}} (\textit{Target Positions Updating}): Assign the new target position $\mathbf{p}_{k+1}[i]$ by solving~\eqref{algo}.
      \vspace{5pt}

  Let $k \leftarrow k+1$, and continue.
   }

   \label{algo:UCB}
\end{algorithm}

\begin{theorem}\label{theorem1}
  Suppose that $\{\mathbf{p}_{k}\}_{ k \in \mathbb{N}_+}$ is the sequence generated by Algorithm~\ref{algo:UCB} {under the conditions in Proposition~\ref{lemma:DUCB}}, then it holds that, with probability $1-\delta$, for $\forall K > 0$,
  \begin{align}
    \sum_{k=1}^K\Big(F_k(\mathbf{p}_k^\star) \hspace{-2pt}-\hspace{-2pt} F_k(\mathbf{p}_k)\Big) \le {\mathcal{O}\Big( \sqrt{K}\log(K)\Big)},
    \end{align}
    where the function $F_k(\cdot)$ and the optimal solutions $\mathbf{p}_k^\star$'s are defined in~\eqref{basicDCM}.
\end{theorem}
\vspace{10pt}

\begin{IEEEproof}
  See Appendix~\ref{subsec:appendB}.
\end{IEEEproof}


\section{Simulation}
In this section, we demonstrate the effectiveness of the proposed algorithm, by considering tracking the moving sources in a pollution diffusion field. In fact, such a problem has been broadly studied in the area of robotics; see e.g.,~\cite{atanasov2014joint,bennetts2013towards,jiang2019source,li2014multi,jiang2018plume}. Compared to these existing works, two primary differences in our problem setup are: 1) we deploy multiple robots/agents, rather than a single one, to the target field; and 2) the pollution distribution in the field is assumed to be disturbed by complex streams such that various local extremum spots are present and therefore the gradient based extremum seeking methods may fail in this scenario. A snapshot of the pollution sources tracking mission is shown in Fig.~\ref{fig:tracking}. Our objective here is to enable the individual robots to track as many moving  pollution sources as possible, through the cooperation among the entire team of robots.

\begin{figure}
  \centering
    \includegraphics[width=1.0\linewidth]{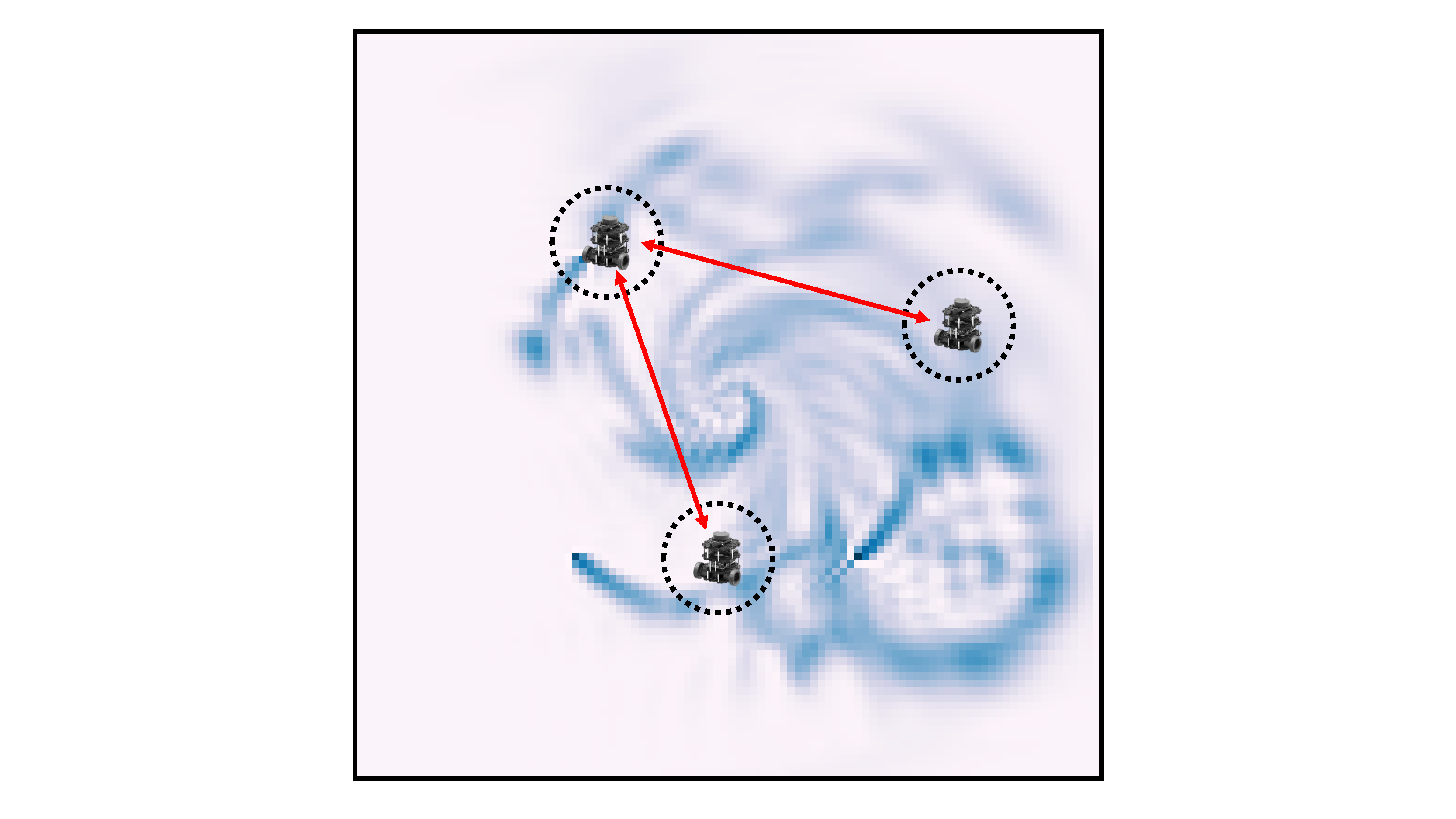}
    \caption{Demonstration of three robots' tracking of the moving sources in an unknown pollution field.}
    \label{fig:tracking}
 \end{figure}

Suppose that the pollution field is described by a~$D \times D$ lattice, as shown in the background of Fig.~\ref{fig:tracking}. Each cell $l \in \{1,2,\cdots, D^2\}$ in the lattice is represented by its position $\mathbf{s}^l$ and also the quantity $\phi_k(\mathbf{s}^l)$ which indicates the pollution level at the discrete time-step $k$. Overall, the $N$-dimensional vector $\bm\phi_k = [\phi_k(\mathbf{s}^1), \phi_k(\mathbf{s}^2),\cdots, \phi_k(\mathbf{s}^N)]^\top$ where $N = D^2$ characterizes the state of the entire pollution field. More specifically, we set {$D = 50$} in this simulation, and consider that the state of field is generated by the discretization of the following convection-diffusion equation~\cite{morton2019revival},
\begin{equation}\label{convection-diff}
  \begin{aligned}
    \frac{\partial \bm\phi_t}{\partial t} = \frac{\lambda}{c\rho}\frac{\partial^2 \bm\phi_t}{\partial^2 x} + \frac{\lambda}{c\rho}\frac{\partial^2 \bm\phi_t}{\partial^2 y} -  \mathbf{u}_x\frac{\partial \bm\phi_t}{\partial x} -  \mathbf{u}_y\frac{\partial \bm\phi_t}{\partial y} + \frac{\mathbf{Q}}{c\rho}.
  \end{aligned}
\end{equation}
Indeed, the similar equation has been widely adopted as a mathematical model {in the study of spread of pollution}; see e.g., \cite{li2014multi,jiang2019source}. Note that here $\mathbf{Q}$ represents the original pollutants, following the diffusion equation as well as the velocity field characterized by $\mathbf{u}_x$ and $\mathbf{u}_y$ in the $x$ and~$y$ directions, respectively. More precisely, we consider that there are three original pollutants in the target field, i.e., $\mathbf{Q} = [Q_1, Q_2, Q_3]$, but the robots have no knowledge about them. Other field related parameters are assumed to be a known prior, so that the Kalman consensus filter can be performed to estimate the unknown states.
In order to track the moving pollution sources, we employ a team of three robots as shown in Fig.~\ref{fig:tracking}, each of them is equipped with a sensor that is capable of measuring a circular area with radius $r = 3$; see the detailed measurement model~\eqref{dynMeasurement} and the description of measurement matrix~\eqref{circularMeasure} in Remark~\ref{remark:measurement}. In particular, we assume that the sensing noise of each robot is independent and identically Gaussian distributed with zero-mean and covariance {$V^i = \mathbf{I}$}, where $\mathbf{I}$ denotes the identity matrix with appropriate dimension. {Note that, since the maximum value of the state $\bm\phi_k$ is set around 5, the noise covariance is reasonably large so that the overall problem is essentially non-trivial to solve.} Besides, it is also assumed that the three robots can exchange information with their immediate neighbors, and the communication channels, shown as the red dot lines in Fig. 2, follow a simple undirected connected graph.


\begin{figure}
\centering
  \begin{subfigure}[b]{1.0\linewidth}
  \centering
    \includegraphics[width=\textwidth]{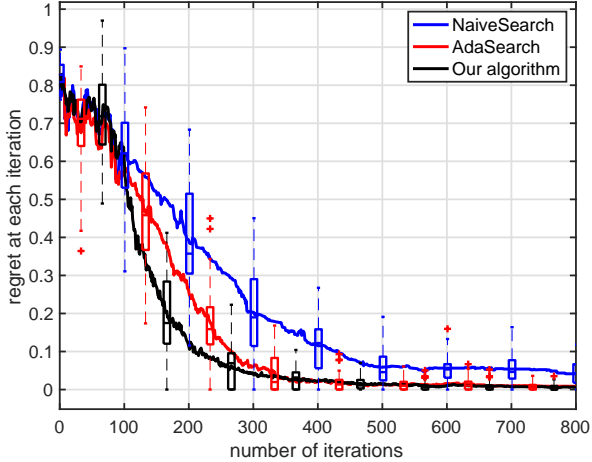}
    \caption{Regret $r^k$ at each iteration $k$}
    \label{fig:1}
  \end{subfigure}
  \begin{subfigure}[b]{1.0\linewidth}
  \centering
    \includegraphics[width=\textwidth]{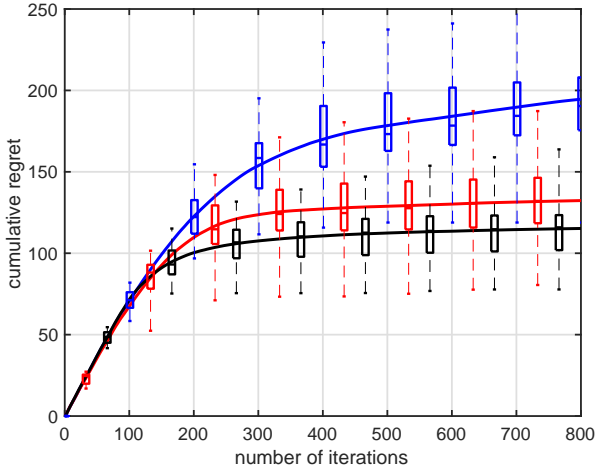}
    \caption{Cumulative regret $\sum_{t=1}^k r^t$}
    \label{fig:2}
  \end{subfigure}
    \caption{Comparison of the regret with three different schemes}
  \label{fig:comparison}
\end{figure}

To demonstrate the result of tracking of the moving pollution sources, Fig.~\ref{fig:1} and Fig.~\ref{fig:2} show the regret defined as $r^k = F_k\big(\mathbf{p}^\star_k\big) - F_k\big(\mathbf{p}_k\big)$ at each iteration~$k$ as well as the cumulative regret defined as $\sum_{t=1}^k r^t$, respectively. Note that while each curve shows the result averaged from {$30$} Monte-Carlo trials, the boxes demonstrate the variance for each independent trial. Further, we also compare the performance of our distributed extreme seeking algorithm with two other existing schemes: 1) the  algorithm proposed in~\cite{rolf2020successive}; and 2) a naive approach, termed as \texttt{NaiveSearch}, in which the robots scan the whole unknown field repeatedly and determine the position of the pollution sources by the current estimation of the field. Notice that in the previous work~\cite{rolf2020successive}, both \texttt{AdaSearch} and \texttt{NaiveSearch} only deal with the static environment with a single robot. In order to compare with them in a fair way, we adopted the same Kalman consensus filter to estimate the unknown dynamical pollution field but apply different searching strategies to seek the pollution sources. It can be concluded from Fig.~\ref{fig:comparison} that the regret $r^k$ generated by our algorithm decreases to zero as the number of iterations grows, which confirms that the team of robots will be able to track the moving pollution sources. In addition, our algorithm achieves the fastest regret descending rate, meaning that the pollution sources will be tracked more efficiently than the two schemes. The cumulative regret shows a sub-linear increase for the proposed algorithm, which is also consistent with the theoretical result presented in Theorem~\ref{theorem1}.

\section{Conclusion}
{In this paper, we proposed a novel algorithmic framework for solving the multi-agent on-line extremum seeking problem in an unknown, dynamical environment. Building on the notion of D-UCB, our algorithm integrates the estimation of the unknown environment and task planning for multiple agents in the on-line manner, and more importantly, significantly reduces the computational complexity of solving the maximization subproblems. Both theoretical analysis and numerical simulations show that our algorithm can enable the network of agents to dynamically track the moving extremum spots presented in the unknown environment. A primary direction of the future works will be focused on the development of algorithm dealing with a more general environment setup; for example, considering the states of environment to be affected by some process noise and/or unknown disturbances.}

\section{Appendix}

In order to facilitate the following proofs, let us start with introducing several vector norms. First, associated with an arbitrary positive definite matrix $M =[m_{ij}]_{i,j=1}^N \in \mathbb{R}^{N \times N}$, we define the $\mathcal{L}_2$-based vector norm $\|\cdot\|_M : \mathbb{R}^N \to \mathbb{R}_+$ as 
\begin{align}\label{2norm}
  \|\mathbf{x}\|_M := \sqrt{\mathbf{x}^\top M\mathbf{x}},
\end{align}
where $\mathbf{x} = [x_1, x_2, \cdots x_N]^\top \in \mathbb{R}^N$. Further, let us define~the $\mathcal{L}_\infty$-based norm $\|\cdot\|_{\mathcal{D}_M, \infty}: \mathbb{R}^N \to \mathbb{R}_+$ associated with the diagonal matrix of the arbitrary positive definite $M$, i.e., {$\mathcal{D}_M = \text{Diag}\{m_{11}, m_{22}, \cdots, m_{NN}\} \in \mathbb{R}^{N \times N}$, as}
\begin{align}\label{Linfty}
  \|\mathbf{x}\|_{\mathcal{D}_M, \infty} := \max_{1 \le i\le N} \; m_{ii}\cdot|x_i|.
\end{align}
Note that the above norm $\|\cdot\|_{\mathcal{D}_M, \infty}$ is well-defined since the positive definiteness of $M$ ensures that $m_{ii} > 0$. Similarly, we define the $\mathcal{L}_1$-based norm \mbox{$\|\cdot\|_{\mathcal{D}_M, 1}: \mathbb{R}^N \to \mathbb{R}_+$} as
\begin{align}\label{L1}
  \|\mathbf{x}\|_{\mathcal{D}_M,1} := \sum_{i=1}^N m_{ii}\cdot|x_i|.
\end{align}

With the vector norms introduced above, it can be immediately verified that the $\mathcal{L}_1$-based norm $\|\cdot\|_{D, 1}$ is the dual norm of the $\mathcal{L}_\infty$-based $\|\cdot\|_{D^{-1}, \infty}$ where $\mathcal{D}_M^{-1}$ takes the inverse of the matrix $\mathcal{D}_M$, and for $\forall \mathbf{x} \in \mathbb{R}^N$,
\begin{align}\label{1to2norm}
  \|\mathbf{x}\|_{\mathcal{D}_M,\infty}\le\|\mathbf{x}\|_{\mathcal{D}_M,1} \le \sqrt{N}\cdot \|\mathbf{x}\|_{\mathcal{D}_M^{2}}.
\end{align}
In addition, we show, by the following lemma, the relationship between $\|\mathbf{x}\|_M$ and $\|\mathbf{x}\|_{\mathcal{D}_M}$.

\begin{lemma}\label{lemma:diagNorm}
  For arbitrary positive definite $M\in \mathbb{R}^{N \times N}$, it holds that $\forall \mathbf{x} \in \mathbb{R}^N$,
  \begin{align}
    \|\mathbf{x}\|_M \le N \cdot \|\mathbf{x}\|_{\mathcal{D}_M}.
  \end{align}
\end{lemma}
\begin{IEEEproof}
  According to the above definitions, one can have that
  \begin{equation}
    \begin{aligned}
      \|\mathbf{x}\|_M^2 &= \sum_{i=1}^N \sum_{j = 1}^N m_{ij}\cdot x_i x_j\\
      & \le \sum_{i=1}^N m_{ii}\cdot x_i^2 + \sum_{i=1}^N \sum_{j\neq i} |m_{ij}|\cdot|x_i x_j|\\
      & \le \sum_{i=1}^N m_{ii}\cdot x_i^2 + \sum_{i=1}^N \sum_{j\neq i} \sqrt{m_{ii}m_{jj}}\cdot|x_i x_j|\\
      & \le \sum_{i=1}^N m_{ii}\cdot x_i^2 + \sum_{i=1}^N \sum_{j\neq i} \frac{1}{2}(m_{ii}\cdot x_i^2 + m_{jj}\cdot x_j^2)\\
      & = N\cdot \sum_{i=1}^N m_{ii}\cdot x_i^2\\
      & = N \cdot \|\mathbf{x}\|_{\mathcal{D}_M}.
    \end{aligned}
  \end{equation}
  Note that the second inequality is due to the positive definiteness of $M$, i.e., $|m_{ij}| \le \sqrt{m_{ii}m_{jj}}$. Therefore, the proof is completed.
\end{IEEEproof}

\subsection{Proof of Proposition~\ref{lemma:DUCB}}\label{subsec:appendA}
By taking advantages of the defined norm $\|\cdot\|_{\mathcal{D}_M, \infty}$, the inequality~\eqref{IneqUCB} in Proposition~\ref{lemma:DUCB} is equivalent to state that, with probability at least $1-\delta$,
\begin{align}\label{IneqUCBinftyNorm}
  &\big\|\widehat{\bm{\phi}}_k - \bm{\phi}_k\big\|_{\mathcal{D}^{-1/2}_{\Sigma_k},\infty} \le \beta_k(\delta).
\end{align}
Therefore, we next prove the inequality~\eqref{IneqUCBinftyNorm} where $\beta_k(\delta)$ is defined in~\eqref{betaK}.

Note that the Kalman consensus filter generates the state estimate $\widehat{\bm{\phi}}_k$ and covariance $\Sigma_k$ as shown in~\eqref{Kalman}, we first show, by the following lemma, an equivalent form of the Kalman consensus filter.

\begin{lemma}\label{lemma:estRecursion}
  Suppose that the state estimates $\widehat{\bm{\phi}}_k$ and covariance $\Sigma_k$ are generated by \eqref{Kalman}, then at each iteration $k$, it is equivalent to write
  \begin{subequations}\label{Kalman_recursion}
    \begin{align}
      \Sigma_k &= A[k:1]\Upsilon_k^{-1}A[k:1]^{\top};\label{Kalman_recursion_Sigma}\\
      \widehat{\bm{\phi}}_k &= A[k:1]\Upsilon_k^{-1}\Big(\Sigma_0^{-1} \widehat{\bm{\phi}}_0 + \sum_{t=0}^{k-1}A[t:1]^\top H_t^\top V^{-1} \mathbf{z}_t\Big),\label{Kalman_recursion_phi}
    \end{align}
   \end{subequations}
   where the matrix $\Upsilon_k \in \mathbb{R}^{N \times N}$ is defined as
  \begin{align}\label{upsilon}
    \Upsilon_k = \Sigma_0^{-1} + \sum_{t=0}^{k-1}A[t:1]^\top H_t^\top V^{-1} H_t A[t:1].
  \end{align}
\end{lemma}

\begin{IEEEproof}
  Let us prove the lemma by mathematical induction. First, it is straightforward to confirm that the above~\eqref{Kalman_recursion} is identical to the original recursion~\eqref{Kalman} when $k=1$. Then, let us assume that \eqref{Kalman_recursion} produces the same results as~\eqref{Kalman} up to the time-step $k$. Next, we prove the consistency for the time-step $k+1$.

  Before proceeding, let us first notice the following identity with the definition of the matrix $\Upsilon_k$,
  \begin{align}\label{identity}
    \Upsilon_{k+1}^{-1} = (\mathbf{I} - \Upsilon_{k+1}^{-1} A[k:1]^\top H_k^\top V^{-1} H_k A[k:1]) \Upsilon_{k}^{-1}.
  \end{align}
  Note that the above equality can be immediately verified by multiplying $\Upsilon_{k+1}$ on the both sides

  Based on the recursion~\eqref{KalmanSigma}, we plug in the previously obtained $\Sigma_{k}$ in the form of~\eqref{Kalman_recursion_Sigma} and have that
  \begin{equation}\label{proof_Sigma}
  \begin{aligned}
    &\Sigma_{k+1} = A_{k+1}\Big(\Sigma_{k}^{-1} + H_k^\top V^{-1} H_k \Big)^{-1}A_{k+1}^\top\\
    &=A_{k+1}\Big(A[k:1]^{-\top}\Upsilon_kA[k:1]^{-1} + H_k^\top V^{-1} H_k \Big)^{-1}A_{k+1}^\top\\
    &=A[k+1:1] \Big(\Upsilon_k + A[k:1]^{\top}H_k^\top V^{-1} H_kA[k:1] \Big)^{-1}\\
    &\hspace{15pt}  \boldsymbol{\cdot}A[k+1:1]^\top\\
    &=A[k+1:1]\Upsilon_{k+1}^{-1}A[k+1:1]^{\top}.
  \end{aligned}
  \end{equation}
  Similarly, we plug $\widehat{\bm{\phi}}_{k}$ in the form of~\eqref{Kalman_recursion_phi} into the~recursion~\eqref{KalmanPhi} and obtain
  \begin{equation}\label{proof_phi}
  \begin{aligned}
    &\widehat{\bm{\phi}}_{k+1} = A_{k+1}\Big(\widehat{\bm{\phi}}_{k} + (\Sigma_{k}^{-1}+Y_k)^{-1}(\mathbf{y}_k - Y_k\widehat{\bm{\phi}}_{k})\Big) \\
    & = A_{k+1}\Big(\mathbf{I} - (\Sigma_{k}^{-1}+Y_k)^{-1}Y_k\Big)\widehat{\bm{\phi}}_{k} \\ 
    & \hspace{10pt}+ A_{k+1}(\Sigma_{k}^{-1}+Y_k)^{-1} H_k^\top V^{-1} \mathbf{z}_k\\
    & = A[k\hspace{-2pt}+\hspace{-2pt}1:1]\Big(\mathbf{I} - \Upsilon_{k+1}^{-1} A[k:1]^\top H_k^\top V^{-1} H_k A[k:1]\Big)\Upsilon_{k}^{-1} \\
    &\hspace{20pt}\boldsymbol{\cdot}\Big(\Sigma_0^{-1} \widehat{\bm{\phi}}_0 + \sum_{t=0}^{k-1}A[t:1]^\top H_t^\top V^{-1} \mathbf{z}_t\Big)\\
    & \hspace{10pt}+ A[k\hspace{-2pt}+\hspace{-2pt}1:1] \Upsilon_{k+1}^{-1} A[k:1]^\top H_k^\top V^{-1} \mathbf{z}_k \\
    & = A[k\hspace{-2pt}+\hspace{-2pt}1:1] \Upsilon_{k+1}^{-1} \Big(\Sigma_0^{-1} \widehat{\bm{\phi}}_0 + \sum_{t=0}^{k}A[t:1]^\top H_t^\top V^{-1} \mathbf{z}_t\Big).
  \end{aligned}
  \end{equation}
  Note that the above identity~\eqref{identity} is applied in the second last~equality. Based on \eqref{proof_Sigma} and~\eqref{proof_phi}, the proof is completed.

\end{IEEEproof}

  
  
  
      
  


Next, given that the state dynamics has $\bm{\phi}_k = A[k:1] \bm{\phi}_0$ and thus $\mathbf{z}_k =H_k A[k:1] \bm{\phi}_0 + \mathbf{n}_k$, the state estimate $\widehat{\bm{\phi}}_k$ can be further expressed as
\begin{equation}
  \begin{aligned}
    \widehat{\bm{\phi}}_k 
    &= A[k:1]\Upsilon_k ^{-1}\Big(\Sigma_0^{-1} \widehat{\bm{\phi}}_0 +\sum_{t=0}^{k-1}A[t:1]^\top H_t^\top V^{-1} \mathbf{n}_t \\
      &\hspace{10pt}+ \Upsilon_k\bm{\phi}_0  - \Sigma_0^{-1}\bm{\phi}_0\Big)\\
      & =\bm{\phi}_k +A[k:1]\Upsilon_k^{-1}\sum_{t=0}^{k-1}A[t:1]^\top H_t^\top V^{-1} \mathbf{n}_t \\
      &\hspace{10pt}+ A[k:1]\Upsilon_k^{-1}\Sigma_0^{-1} ( \widehat{\bm{\phi}}_0 - {\bm{\phi}}_0).
  \end{aligned}
\end{equation}
Therefore, it holds that $\forall \mathbf{x} \in \mathbb{R}^{N}$,
\begin{equation}\label{errorMeasure}
  \begin{aligned}
    &\mathbf{x}^\top(\widehat{\bm{\phi}}_k - \bm{\phi}_k)\\
      & = \mathbf{x}^\top A[k:1]\Upsilon_k^{-1}\sum_{t=0}^{k-1}A[t:1]^\top H_t^\top V^{-1} \mathbf{n}_t \\
      &\hspace{10pt}+ \mathbf{x}^\top A[k:1]\Upsilon_k^{-1}\Sigma_0^{-1} ( \widehat{\bm{\phi}}_0 - {\bm{\phi}}_0)\\
      &\hspace{-5pt}\overset{(1.a)}{\le} \big\|A[k:1]^\top \mathbf{x}\big\|_{\Upsilon_k^{-1}} \cdot \Big\|\sum_{t=0}^{k-1}A[t:1]^\top H_t^\top V^{-1} \mathbf{n}_t\Big\|_{\Upsilon_k^{-1}}\\
      &\hspace{10pt}+ \big\|A[k:1]^\top \mathbf{x}\big\|_{\Upsilon_k^{-1}} \cdot \big\|\Sigma_0^{-1} ( \widehat{\bm{\phi}}_0 - {\bm{\phi}}_0)\big\|_{\Upsilon_k^{-1}}\\
      & \hspace{-5pt}\overset{(1.b)}{=} \big\|\mathbf{x}\big\|_{\Sigma_k} \cdot\Big(\Big\|\sum_{t=0}^{k-1}A[t:1]^\top H_t^\top V^{-1} \mathbf{n}_t\Big\|_{\Upsilon_k^{-1}}\\
      &\hspace{10pt}+\big\|\Sigma_0^{-1} ( \widehat{\bm{\phi}}_0 - {\bm{\phi}}_0)\big\|_{\Upsilon_k^{-1}}\Big)\\
      &\hspace{-5pt}\overset{(1.c)}{\le} N \cdot \big\|\mathbf{x}\big\|_{\mathcal{D}_{\Sigma_k}} \cdot\Big(\Big\|\sum_{t=0}^{k-1}A[t:1]^\top H_t^\top V^{-1} \mathbf{n}_t\Big\|_{\Upsilon_k^{-1}}\\
      &\hspace{10pt}+\big\|\Sigma_0^{-1} ( \widehat{\bm{\phi}}_0 - {\bm{\phi}}_0)\big\|_{\Upsilon_k^{-1}}\Big).
  \end{aligned}
\end{equation}
where $(1.a)$ is due to the Cauchy-Schwartz inequality; $(1.b)$ is due to \eqref{Kalman_recursion_Sigma}; and $(1.c)$ is based on Lemma~\ref{lemma:diagNorm}.

Now, let $\mathbf{x} = \mathcal{D}^{-1}_{\Sigma_k}(\widehat{\bm{\phi}}_k - \bm{\phi}_k)$, it follows that
\begin{equation}
  \begin{aligned}
    &\big\|\widehat{\bm{\phi}}_k - \bm{\phi}_k\big\|_{\mathcal{D}^{-1}_{\Sigma_k}} \le N \cdot \Big(\Big\|\sum_{t=0}^{k-1}A[t:1]^\top H_t^\top V^{-1} \mathbf{n}_t\Big\|_{\Upsilon_k^{-1}}\\
      &\hspace{100pt}+\big\|\Sigma_0^{-1} ( \widehat{\bm{\phi}}_0 - {\bm{\phi}}_0)\big\|_{\Upsilon_k^{-1}}\Big).
  \end{aligned}
\end{equation}
According to the inequality in~\eqref{1to2norm}, we can have that
\begin{equation}\label{diagError}
  \begin{aligned}
    &\big\|\widehat{\bm{\phi}}_k - \bm{\phi}_k\big\|_{\mathcal{D}^{-1/2}_{\Sigma_k},\infty} \\
    & \le \sqrt{N} \cdot \big\|\widehat{\bm{\phi}}_k - \bm{\phi}_k\big\|_{\mathcal{D}^{-1}_{\Sigma_k}}\\
    &\le N^{3/2} \hspace{-2pt}\cdot\hspace{-2pt} \Big(\Big\|\sum_{t=0}^{k-1}A[t:1]^\top H_t^\top V^{-1} \mathbf{n}_t\Big\|_{\Upsilon_k^{-1}}\\
      &\hspace{45pt}+\big\|\Sigma_0^{-1} ( \widehat{\bm{\phi}}_0 - {\bm{\phi}}_0)\big\|_{\Upsilon_k^{-1}}\Big).
  \end{aligned}
\end{equation}

In order to prove the inequality~\eqref{IneqUCBinftyNorm}, we now need to upper bound the two terms on the right hand side of~\eqref{diagError}; see the following two lemmas.

\begin{lemma}\label{lemma:upperBoundII}
  Let the conditions in Proposition~\ref{lemma:DUCB} hold and the matrix $\Upsilon_k$ be defined as~\eqref{upsilon}, then there exists a constant $C_1 = \|\widehat{\bm{\phi}}_0 - \bm{\phi}_0\| /\sqrt{\ubar{\sigma}}$ such that for $\forall k > 0$,
  \begin{align}
    \big\|\Sigma_0^{-1} ( \widehat{\bm{\phi}}_0 - {\bm{\phi}}_0)\big\|_{\Upsilon_k^{-1}} \le C_1.
  \end{align}
\end{lemma}

\begin{IEEEproof}
  By the definition~\eqref{upsilon} of the matrix $\Upsilon_k$, it is straightforward to see that ${\Upsilon_k^{-1}} \le \Sigma_0$, and therefore,
  \begin{equation}
      \begin{aligned}
    &\big\|\Sigma_0^{-1} ( \widehat{\bm{\phi}}_0 - {\bm{\phi}}_0)\big\|^2_{\Upsilon_k^{-1}}\\ &=  ( \widehat{\bm{\phi}}_0 - {\bm{\phi}}_0)^\top\Sigma_0^{-1}\Upsilon_k^{-1}\Sigma_0^{-1} ( \widehat{\bm{\phi}}_0 - {\bm{\phi}}_0)\\
    & \le ( \widehat{\bm{\phi}}_0 - {\bm{\phi}}_0)^\top\Sigma_0^{-1} ( \widehat{\bm{\phi}}_0 - {\bm{\phi}}_0)\\
    & \le 1/\ubar{\sigma} \cdot \|\widehat{\bm{\phi}}_0 - \bm{\phi}_0\|^2,
  \end{aligned}
  \end{equation}
  where the last inequality is due to the condition $\Sigma_0 \ge \ubar{\sigma} \cdot \mathbf{I}$. Thus, the proof is completed.
\end{IEEEproof}

\begin{lemma}\label{lemma:upperBoundI}
  Let the conditions in Proposition~\ref{lemma:DUCB} hold and the matrix $\Upsilon_k$ be defined as~\eqref{upsilon}, then there exists a constant $C_2' = \bar{v}^2 \sqrt{2N\cdot\max\{1,1/\ubar{v}\}}$ such that with probability at least $1-\delta$, for $\forall k >0$,
  \begin{equation}
  \begin{aligned}
    &\Big\|\sum_{t=0}^{k-1}A[t:1]^\top H_t^\top V^{-1} \mathbf{n}_t\Big\|_{\Upsilon_k^{-1}} \\
    &\hspace{50pt}\le C_2' \cdot \sqrt{\log\Big(\frac{\bar{\sigma}/\ubar{\sigma} + \bar{\alpha}\bar{\sigma}\cdot k / \ubar{v}^2}{\delta^{2/N}}\Big)}.
  \end{aligned}
  \end{equation}

  \begin{IEEEproof}
    This proof is primarily based on the existing results presented in~\cite{abbasi2011improved} (see Lemmas 8 -- 10 and Theorem~1). For the notational simplicity, let us define
    \begin{align}
      X_t := A[t:1]^\top H_t^\top V^{-1} \in \mathbb{R}^{N \times M}.
    \end{align}
    Then, according to Theorem~1 in \cite{abbasi2011improved}, it holds with probability at least $1-\delta$ that,
    \begin{align}\label{upperBoundI}
      \Big\|\sum_{t=0}^{k-1} X_t\mathbf{n}_t\Big\|_{\Omega_k^{-1}} \le 2\bar{v}^2 \cdot\sqrt{\log\Big(\frac{\det(\Omega_k)^{1/2}\det(\Sigma_0)^{1/2}}{\delta}\Big)},
    \end{align}
    where $\Omega_k := \Sigma_0^{-1} + \sum_{t=0}^{k-1} X_t X_t^\top\in\mathbb{R}^{N \times N}$. Let us recall the definition~\eqref{upsilon} of the matrix $\Upsilon_k$ and notice that there is a slight difference between $\Omega_k$ and $\Upsilon_k$. Next, we show that there exists a constant $C'_3  = \max\{1, 1/\ubar{v}\}$ such that $\Omega_k \le C'_3 \cdot\Upsilon_k, \forall k>0$. In fact, it holds that
    \begin{equation}
    \begin{aligned}
      \Omega_k &= \Sigma_0^{-1} +  \sum_{t=0}^{k-1}A[t:1]^\top H_t^\top V^{-2} H_t A[t:1] \\
      &\le \Sigma_0^{-1} + 1/\ubar{v}\cdot \sum_{t=0}^{k-1}A[t:1]^\top H_t^\top V^{-1} H_t A[t:1] \\
      &\le \max\{1, 1/\ubar{v}\} \cdot \Upsilon_k.
    \end{aligned}
    \end{equation}
    Note that the first inequality is due to the fact that $\ubar{v}$ is the smallest entry of the diagonal matrix $V$; see Assumption~\ref{assump:measurementNoise}. Therefore, the previous statement can be immediately proved by letting $C'_3 = \max\{1, 1/\ubar{v}\}$. Now, based on such a statement, it holds that $\Upsilon_k^{-1} \le C'_3\cdot \Omega_k^{-1}$. Together with the inequality \eqref{upperBoundI}, one can have that
    \begin{equation}
      \begin{aligned}
         &\Big\|\sum_{t=0}^{k-1} X_t\mathbf{n}_t\Big\|_{\Upsilon_k^{-1}} \le \sqrt{C'_3} \cdot \Big\|\sum_{t=0}^{k-1} X_t\mathbf{n}_t\Big\|_{\Omega_k^{-1}}\\
         &\le 2\bar{v}^2\sqrt{\max\{1, 1/\ubar{v}\}} \cdot\hspace{-2pt}\sqrt{\log\Big(\frac{\det(\Omega_k)^{1/2}\det(\Sigma_0)^{1/2}}{\delta}\Big)}.
      \end{aligned}
    \end{equation}

   Moreover, according to the inequality of arithmetic and geometric means and the definition of $\Omega_k$, it holds that
    \begin{equation}
      \begin{aligned}
        \det(\Omega_k) &\le \Big(1/N\cdot\text{trace}\big(\Sigma_0^{-1}\big)+1/N\cdot\sum_{t=0}^{k-1} \text{trace}(X_tX_t^\top)\Big)^N,
      \end{aligned}
    \end{equation}
    where the trace of the matrix $X_t X_t^\top$ further has
    \begin{equation}
      \begin{aligned}
        \text{trace}(X_tX_t^\top) &= \text{trace}\Big(A[t:1]^\top H_t^\top V^{-2} H_t A[t:1]\Big)\\
        &\hspace{-5pt}\overset{(2.a)}{\le} 1/\ubar{v}^2 \cdot \sum_{n=1}^N \mathbf{e}_n^\top A[t:1]^\top H_t^\top H_t A[t:1]\mathbf{e}_n\\
        &\hspace{-5pt}\overset{(2.b)}{\le} 1/\ubar{v}^2 \cdot \sum_{n=1}^N \mathbf{e}_n^\top A[t:1]^\top A[t:1]\mathbf{e}_n\\
        &\hspace{-5pt}\overset{(2.c)}{\le} N\cdot\bar{\alpha}/\ubar{v}^2.
      \end{aligned}
    \end{equation}
    Note that $(2.a)$ is due to Assumption~\ref{assump:measurementNoise} and $\mathbf{e}_n \in \mathbb{R}^N$ denotes the unit vector; $(2.b)$ follows from the special form of the measurement matrix $H_t$, i.e., each row has only one element equal to one and all others equal to zero; and $(2.c)$ is based on Assumption~\ref{assump:dynamicsBound}. In addition, given that the initialization $\Sigma_0$ ensures $\ubar{\sigma} \cdot \mathbf{I} \le \Sigma_0 \le \bar{\sigma} \cdot \mathbf{I}$, it follows that $\text{trace}(\Sigma_0^{-1}) \le N/\ubar{\sigma}$ and $\det(\Sigma_0) \le \bar{\sigma}^N$. As a result, we can eventually arrive at
    \begin{equation}
    \begin{aligned}
      &\sqrt{\log\Big({\det(\Omega_k)^{1/2}\det(\Sigma_0)^{1/2}}/{\delta}\Big)} \\
      &=\sqrt{ 1/2\cdot\log\big(\det(\Omega_k)\big) + 1/2\cdot \log\big(\det(\Sigma_0)\big) - \log(\delta)}\\
      &\le \sqrt{N/2}\cdot\sqrt{\log\Big(\frac{\bar{\sigma}/\ubar{\sigma} + \bar{\alpha}\bar{\sigma}\cdot k / \ubar{v}^2}{\delta^{2/N}}\Big)}.
    \end{aligned}
    \end{equation}
    Together with the inequality~\eqref{upperBoundI}, the proof of Lemma~\ref{lemma:upperBoundI} is completed.

  \end{IEEEproof}
\end{lemma}

Now, based on Lemmas~\ref{lemma:upperBoundII} -- \ref{lemma:upperBoundI} and inequality~\eqref{diagError}, it has been shown that, {with probability $1 - \delta$}
\begin{equation}
  \begin{aligned}
  &\big\|\widehat{\bm{\phi}}_k - \bm{\phi}_k\big\|_{\mathcal{D}^{-1/2}_{\Sigma_k},\infty} \\
  &\le N^{3/2} \cdot \Bigg(C_1 + C'_2\cdot \sqrt{\log\Big(\frac{\bar{\sigma}/\ubar{\sigma} + \bar{\alpha}\bar{\sigma}\cdot k / \ubar{v}^2}{\delta^{2/N}}\Big)}\Bigg),
\end{aligned}
\end{equation}
with $C_1 =\hspace{-2pt} \|\widehat{\bm{\phi}}_0 - \bm{\phi}_0\|^2 /\ubar{\sigma}$ and $C_2' =\hspace{-2pt}  \bar{v}^2\sqrt{2N\cdot\max\{1,1/\ubar{v}\}}$. Therefore, Proposition~\ref{lemma:DUCB} is proved.

\subsection{Proof of Theorem~\ref{theorem1}}\label{subsec:appendB}

Let us start the proof by introducing additional notations. Recall that $\mathbf{p}_k^\star$, as defined in~\eqref{basicDCM}, denotes the positions of the moving extremum spots at time-step $k$, and similarly, $\mathbf{p}_k$ denotes the target positions for the multiple agents generated by our algorithm. To better characterize the positional information, let us define a mapping $\mathbf{a}(\cdot): \mathcal{S}^I \to \mathbb{R}^N$ which maps the position $\mathbf{p}$ to the $N$-dimensional vector,
\begin{align}\label{defAction}
  \mathbf{a}(\mathbf{p}) = \sum_{i=1}^I \mathbf{e}_{s_i},
\end{align}
where each $s_i$ corresponds to the index of the position~$\mathbf{p}[i]$. More precisely, since the positions $\mathbf{p}_k$ and $\mathbf{p}^\star_k$ are solved by the maximization problems; see~\eqref{algo} and \eqref{basicDCM}, it can be immediately verified that the vectors $\mathbf{a}(\mathbf{p}_k)$ and $\mathbf{a}(\mathbf{p}^\star_k)$ must have $I$ elements equal to one and all others equal to zero. Therefore, we denote the set of all possibilities of these vectors as
\begin{align}
  \mathcal{A} := \{\mathbf{a} \, |\, \mathbf{a} \in \{0,1\}^N, \mathbf{1}^\top \mathbf{a} = I\}.
\end{align}
Furthermore, for the notational simplicity, we abbreviate the above $\mathbf{a}(\mathbf{p}_k)$ and $\mathbf{a}(\mathbf{p}^\star_k)$ to $\mathbf{a}_k \in \mathcal{A}$ and $\mathbf{a}^\star_k \in \mathcal{A}$, respectively. With the help of these notations, the loss of function values can be expressed as,
\begin{align}\label{lossValue}
  {r_k}: = F_k\big(\mathbf{p}^\star_k\big) - F_k(\mathbf{p}_k) = \langle \mathbf{a}^\star_k - \mathbf{a}_k, \bm{\phi}_k\rangle.
\end{align}

Next, we show, by the following lemma, that there exists an uniform upper bound for the loss of function values.
\begin{lemma}\label{lemma:lossUpperBound}
  Suppose that Assumption~\ref{assump:dynamicsBound} holds and the loss of function $r_k$ is defined as~\eqref{lossValue}, then there is an upper bound \mbox{$\bar{\gamma} = 2\sqrt{I \bar{\alpha}} \cdot\|\bm{\phi}_0\|^2$} such that for $r_k \le \bar{\gamma},\,\forall k >0$.
\end{lemma}
\begin{IEEEproof}
  Recall that the linear dynamics of the state $\bm{\phi}_k$ ensures $\bm{\phi}_k = A[k:1] \bm{\phi}_0$, thus based on~\eqref{lossValue}, it follows that
  \begin{equation}
    \begin{aligned}
      r_k &\hspace{-5pt}\overset{(3.a)}{\le} \|\mathbf{a}_k^\star - \mathbf{a}_k\|\cdot \|\bm{\phi}_k\|\\
      &\hspace{-5pt}\overset{(3.b)}{\le} \big(\|\mathbf{a}_k^\star\| + \|\mathbf{a}_k\|\big) \cdot \big\| \bm{\phi}_0^\top A[k:1]^\top  A[k:1] \bm{\phi}_0\big\| \\
      & \hspace{-5pt}\overset{(3.c)}{\le}2\sqrt{I \bar{\alpha}} \cdot\|\bm{\phi}_0\|^2
    \end{aligned}
  \end{equation}
  where $(3.a)$ is due to the Cauchy-Schwartz inequality; $(3.b)$ follows from the triangle inequality and the state dynamics; and $(3.c)$ is based on the fact that both $\mathbf{a}_k^\star $ and $ \mathbf{a}_k$ are from the set $\mathcal{A}$ as well as Assumption~\ref{assump:dynamicsBound}.
\end{IEEEproof}

Let us define another set $\bm{\chi}^k \in \mathbb{R}^N$ which is~characterized by Proposition~\ref{lemma:DUCB} (or the inequality~\eqref{IneqUCBinftyNorm}),
\begin{align}\label{polytope}
  \bm{\chi}^k: = \big\{\bm{\phi} \, \big\vert\, \|\widehat{\bm{\phi}}_k - \bm{\phi}\|_{\mathcal{D}^{-1/2}_{\Sigma_k},\infty} \le \beta_k(\delta)\big\}.
\end{align}
It is guaranteed by Proposition~\ref{lemma:DUCB} that $\bm{\phi}_k$ must be in the set $\bm{\chi}^k$ with probability at least $1-\delta$ at each time-step $k$. 

With the help of the defined set $\bm{\chi}^k$, we now present a supporting lemma which measures the update of the target positions~$\mathbf{p}_k$~(or $\mathbf{a}_k$) at each time-step $k$.




\begin{lemma}\label{lemma:maxValue}
  Under the conditions in Proposition~\ref{lemma:DUCB}, suppose that the positional information $\mathbf{a}_k$ is generated by solving the maximization problem~\eqref{algo} with the D-UCB $\bm{\mu}_k$ computed by~\eqref{UCB}, then the optimal function value $\langle \mathbf{a}_k, \bm{\mu}_k\rangle$ of~\eqref{algo} can be obtained by solving the following constrained bi-linear program,
  \begin{align}\label{bilinearProb}
    \mathop {\text{maximize} }\limits_{\mathbf{a} \in \mathcal{A}, \bm{\phi} \in \bm{\chi}^k}  \quad \langle \mathbf{a},\; \bm{\phi}\rangle.
  \end{align}
  In addition, it holds with probability $1-\delta$ that,
  \begin{align}\label{optValueIneq}
    \langle \mathbf{a}_k,\; \bm{\mu}_k\rangle \ge \langle \mathbf{a}_k^\star,\; \bm{\phi}_k\rangle.
  \end{align}
\end{lemma}

\begin{IEEEproof}
  Notice that the constraint bi-linear problem~\eqref{bilinearProb} can be written as the following equivalent form,
  \begin{align}
    \mathop {\text{maximize} }\limits_{\mathbf{a} \in \mathcal{A}}  \quad Q(\mathbf{a}),
  \end{align}
  where the objective function $Q(\cdot): \mathcal{A} \to \mathbb{R}$ is defined by another maximization problem,
  \begin{align}\label{equavilentMax}
    Q(\mathbf{a}):= \mathop {\text{maximize} }\limits_{ \bm{\phi} \in \bm{\chi}^k}  \quad \langle \mathbf{a},\; \bm{\phi}\rangle.
  \end{align}

  Based on the KKT conditions and the definition of the feasible set $\bm{\chi}^k$, the optimal solution $\bm{\phi}^\star$ of the problem~\eqref{equavilentMax} can be analytically expressed as
  \begin{align}
    \bm{\phi}^\star = \widehat{\bm{\phi}}_k + \beta_k(\delta)\cdot \text{diag}^{1/2}(\Sigma_k),
  \end{align}
  which is exactly the same as {the definition of D-UCB in~\eqref{UCB}}. Therefore, it holds that
  \begin{align}
     \langle \mathbf{a}_k,\; \bm{\mu}_k\rangle  = \mathop {\text{maximize} }\limits_{\mathbf{a} \in \mathcal{A}, \bm{\phi} \in \bm{\chi}^k}  \quad \langle \mathbf{a},\; \bm{\phi}\rangle.
  \end{align}

  Furthermore, since Proposition~\ref{lemma:DUCB} guarantees that $\bm{\phi}_k \in \bm{\chi}^k$ with probability $1-\delta$ and $\mathbf{a}^\star = \argmax_{\mathbf{a} \in \mathcal{A}} \langle \mathbf{a},\,\bm{\phi}_k \rangle$, it is straightforward to verify that the inequality~\eqref{optValueIneq} holds with probability $1-\delta$.
\end{IEEEproof}

Now, we are ready to prove the statement in Theorem~\ref{theorem1}, i.e., $\sum_{k = 1}^K r_k \le \mathcal{O}\big(\sqrt{K} \log{(K)}\big)$. Before proceeding, let us first recall that the vector norm $\|\cdot\|_{\mathcal{D}_M, 1}$ as defined in~\eqref{L1} is the dual norm of $\|\cdot\|_{\mathcal{D}_M^{-1}, \infty}$ as defined in~\eqref{Linfty}. Therefore, the loss of function value $r_k$ has
\begin{equation}\label{lossIneq}
  \begin{aligned}
    r_k & =  \langle \mathbf{a}^\star_k, \; \bm{\phi}_k\rangle - \langle \mathbf{a}_k, \; \bm{\phi}_k\rangle \\
    &\hspace{-5pt}\overset{(4.a)}{\le} \langle \mathbf{a}_k, \; \bm{\mu}_k\rangle - \langle \mathbf{a}_k, \; \bm{\phi}_k\rangle \\
    & = \langle \mathbf{a}_k, \; \bm{\mu}_k -  \bm{\phi}_k\rangle \\
    &\hspace{-5pt}\overset{(4.b)}{\le} \|\mathbf{a}_k\|_{\mathcal{D}^{1/2}_{\Sigma_{k}},1} \cdot \|\bm{\mu}_k -  \bm{\phi}_k\|_{\mathcal{D}^{-1/2}_{\Sigma_{k}},\infty}\\
    &\hspace{-5pt}\overset{(4.c)}{\le}2\beta_k(\delta)\cdot \|\mathbf{a}_k\|_{\mathcal{D}^{1/2}_{\Sigma_{k}},1}\\
    &\hspace{-5pt}\overset{(4.d)}{\le}2\sqrt{N}\beta_k(\delta)\cdot \|\mathbf{a}_k\|_{\mathcal{D}_{\Sigma_{k}}},
  \end{aligned}
\end{equation}
  where the inequality $(4.a)$ is due to the above Lemma~\ref{lemma:maxValue}; $(4.b)$ follows from the H\"older's inequality; $(4.c)$ is due to the triangle inequality and the fact that both $\bm{\mu}_k$ and $\bm{\phi}_k$ are in the set $\bm{\chi}^k$; and $(4.d)$ comes from the inequality \eqref{1to2norm}. Next, in order to further investigate the key term $\|\mathbf{a}_k\|_{\mathcal{D}_{\Sigma_{k-1}}}$, we show an upper bound for the cumulative $\|\mathbf{a}_k\|_{\mathcal{D}_{\Sigma_{k-1}}}$'s with respect to the time-step $k$. 

\begin{lemma}\label{lemma:normBound}
  Suppose that the conditions in Proposition~\ref{lemma:DUCB} hold and the positional information $\mathbf{a}_k$'s are generated by Algorithm~\ref{algo:UCB}, then it holds that for $\forall K >0$,
  \begin{equation}
  \begin{aligned}
  &\sum_{k=0}^{K-1} \min\{1, 1/\bar{v}\cdot \|\mathbf{a}_k\|^2_{\mathcal{D}_{\Sigma_{k}}}\}\\
  &\le 2N \cdot \log\Big(\det(\Sigma_0)^{1/N} \cdot \bar{\alpha} \cdot\big((\ubar{\alpha}\ubar{\sigma})^{-1} + K\cdot(\ubar{\alpha} \ubar{v})^{-1}\big)\Big).
  \end{aligned}
  \end{equation}
\end{lemma}
\begin{IEEEproof}
  Recall that the matrix $\Sigma_k$ is generated by the following recursion,
  \begin{align}
    \Sigma_{k+1} = A_{k+1}\Big(\Sigma_{k}^{-1} + H_k^\top V^{-1} H_k \Big)^{-1}A_{k+1}^\top.
  \end{align}
  For the sake of presentation, let us first focus on the inverse of $\Sigma_k$, i.e., $\Theta_k = \Sigma_k^{-1} \in \mathbb{R}^{N \times N}$, and thus it holds that,
  \begin{align}\label{ThetaRecursion}
    \Theta_{k+1} = A_{k+1}^{-\top}\big(\Theta_{k}+H_k^\top V^{-1} H_k\big)A_{k+1}^{-1}  .
  \end{align}

  Consider the determinant of the matrices $\Theta_k$'s, then one can have that
  \begin{equation}\label{det}
    \begin{aligned}
      &\det(\Theta_{k+1})\\
      &= 1/\det(A_{k+1}^\top A_{k+1})\cdot \det\big(\Theta_{k}+H_k^\top V^{-1} H_k\big)\\
      &=1/\det(A_{k+1}^\top A_{k+1})  \\
      &\hspace{15pt}\boldsymbol{\cdot}\det\Big(\Theta_{k}^{1/2}\big(\mathbf{I}+ \Theta_{k}^{-1/2}H_k^\top V^{-1} H_k\Theta_{k}^{-1/2}\big)\Theta_{k}^{1/2}\Big)\\
      &=\det(\Theta_{k})/\det(A_{k+1}^\top A_{k+1}) \\
      &\hspace{12pt} \boldsymbol{\cdot}\det\Big(\mathbf{I} +\Theta_{k}^{-1/2} H_k^\top V^{-1} H_k\Theta_{k}^{-1/2}\Big).
    \end{aligned}
  \end{equation}
  For simplicity, we here use $Y_k$ to substitute $H_k^\top V^{-1} H_k$ again. Consider that the noise covariance matrix $V$ is diagonal and $H_k$ takes the specific form of 
  \begin{align}
    H_k = [\mathbf{e}_l]_{l \in \cup_{i=1}^I \mathcal{C}^i}^\top,
  \end{align}
  where each set $\mathcal{C}^i$ contains the indices of the positions covered by the agent $i$'s sensing area. Therefore, the matrix $Y_k$ is also diagonal and can be expressed as 
  \begin{align}
    Y_k = \sum_{i=1}^I  \sum_{l \in \mathcal{C}^i} 1/v^i \cdot\mathbf{e}_l\mathbf{e}_l^\top.
  \end{align}
  Further, let us denote $\Theta_{k}^{-1/2} Y_k\Theta_{k}^{-1/2}$ by $\Xi_k \in \mathbb{R}^{N \times N}$. Suppose that $\lambda_n(\Xi_k)$ represents the $n$-th eigenvalue and $\xi_{nn}^k$ is the $n$-th diagonal entry of $\Xi_k$, then the trace of the matrix has
  \begin{align}\label{trace}
    \text{trace}({\Xi_k}) = \sum_{n=1}^N \lambda_n(\Xi_k) = \sum_{n=1}^N \xi_{nn}^k.
  \end{align}
  In addition, we denote $\bm{\theta}^{k}_n \in \mathbb{R}^N$ the $n$-th column of the matrix $\Theta_{k}^{-1/2}$; note that $(\bm{\theta}^{k}_n)^\top$ is also the $n$-th row since $\Theta_{k}^{-1/2}$ is symmetric. As a result of the specific structure of the matrix $Y_k$, the diagonal entries $\xi_{nn}^k$ of $\Xi_k$ has
  \begin{equation}\label{diagPsi}
    \begin{aligned}
      \xi_{nn}^k &\hspace{-5pt}\overset{(5.a)}{=} \Big(\sum_{i=1}^I \delta^i_k(n)/v^i\Big)\cdot(\bm{\theta}^{k}_n)^\top\bm{\theta}^{k}_n\\
      &\hspace{-5pt}\overset{(5.b)}{=} \Big(\sum_{i=1}^I \delta^i_k(n)/v^i\Big)\sigma_{nn}^{k}\\
      &\hspace{-5pt}\overset{(5.c)}{\ge} 1/\bar{v}\cdot\sum_{i=1}^I \delta^i_k(n)\sigma_{nn}^{k},
    \end{aligned}
  \end{equation}
  where in  $(5.a)$, we let $\delta^i_k(n) = 1$ if the position indexed by~$n$ is in the sensing area $\mathcal{C}^i$ at the time-step $k$, and $\delta^i_n = 0$ otherwise;
  $(5.b)$ is due to the definition of $\bm{\theta}^{k}_n$ and the fact that $ \sigma_{nn}^{k}$ denotes the $n$-th diagonal entry of $\Sigma_{k}$; and $(5.c)$ comes from Assumption~\ref{assump:measurementNoise}.
  Now, based on~\eqref{diagPsi}, one can further have that
  \begin{equation}\label{diagIneq}
    \begin{aligned}
      \sum_{n =1}^N  \xi_{nn}^k &\ge 1/\bar{v}\cdot \sum_{n=1}^N \sum_{i=1}^I \delta^i_k(n)\sigma_{nn}^{k}\\
      &\hspace{-5pt}\overset{(6.a)}{\ge} 1/\bar{v}\cdot \sum_{i=1}^I \mathbf{e}_{s^i_k}^\top\Sigma_{k}\mathbf{e}_{s^i_k}\\
      &\hspace{-5pt}\overset{(6.b)}{=} 1/\bar{v}\cdot \mathbf{a}_k^\top \mathcal{D}_{\Sigma_{k}} \mathbf{a}_k\\
      &\hspace{-5pt}\overset{(6.c)}{=}1/\bar{v}\cdot \|\mathbf{a}_k\|^2_{\mathcal{D}_{\Sigma_{k}}},
    \end{aligned}
  \end{equation}
  where $s^i_k$ denotes the index of the agent $i$'s position at the time-step $k$ in $(6.a)$ and $\delta_k^i (s_k^i)$ must be one; $(6.b)$ is by the definition \eqref{defAction} of $\mathbf{a}_k$ and $(6.c)$ is due to the definition of the norm $\|\cdot\|_{\mathcal{D}_{\Sigma_{k}}}$.

  Now, the previous equalities in \eqref{det} can be continued as
    \begin{equation}\label{detIneq}
    \begin{aligned}
      &\det(\Theta_{k+1}) =\det(\Theta_{k})/\det(A_{k+1}^\top A_{k+1})\cdot\det(\mathbf{I} +\Xi_k)\\
      &\hspace{-5pt}\overset{(7.a)}{=}\det(\Theta_{k})/\det(A_{k+1}^\top A_{k+1})\cdot \prod_{n=1}^N \big(1 + \lambda_n(\Xi_k)\big)\\
      &\hspace{-5pt}\overset{(7.b)}{\ge}\det(\Theta_{k})/\det(A_{k+1}^\top A_{k+1})\cdot \Big(1 + \sum_{n=1}^N \lambda_n(\Xi_k)\Big)\\
      &\hspace{-5pt}\overset{(7.c)}{=}\det(\Theta_{k})/\det(A_{k+1}^\top A_{k+1})\cdot\Big(1 + \sum_{n=1}^N \xi_{nn}^k\Big)\\
      &\hspace{-5pt}\overset{(7.d)}{\ge}\det(\Theta_{k})/\det(A_{k+1}^\top A_{k+1})\cdot \Big(1 +  1/\bar{v}\cdot \|\mathbf{a}_k\|^2_{\mathcal{D}_{\Sigma_{k}}}\Big),
    \end{aligned}
  \end{equation}
  where $(7.a)$ is due to the fact that the determinant of a matrix equals the product of eigenvalues; $(7.b)$ follows from the inequality of arithmetic and geometric means and the positive definiteness of the matrix $\Xi_k$; $(7.c)$ is based on the equality~\eqref{trace}; and $(7.d)$ is due to the inequality~\eqref{diagIneq}. Subsequently, applying~\eqref{detIneq} recursively yields
  \begin{equation}
  \begin{aligned}
    \det(\Theta_{k+1}) &\ge \det(\Theta_0)/\det\big(A[k+1:1]^\top A[k+1:1]\big) \\
    &\hspace{15pt}\boldsymbol{\cdot}\prod_{t=0}^k \Big(1 +  1/\bar{v}\cdot \|\mathbf{a}_t\|^2_{\mathcal{D}_{\Sigma^{t}}}\Big)\\
    &\ge \bar{\alpha}^{-N} \det(\Theta_0)\cdot \prod_{t=0}^k \Big(1 +  1/\bar{v}\cdot \|\mathbf{a}_t\|^2_{\mathcal{D}_{\Sigma^{t}}}\Big).
  \end{aligned}
  \end{equation}
  Note that the last inequality relies on Assumption~\ref{assump:dynamicsBound}.

  Next, notice that $\min\{1,x\} \le 2\log(1+x)$ is always true for any non-negative scalar $x\ge 0$, therefore,
  \begin{equation}\label{minIneq}
    \begin{aligned}
      &\sum_{t = 0}^k \min\{1, 1/\bar{v} \cdot\|\mathbf{a}_t\|^2_{\mathcal{D}_{\Sigma^{t}}}\}\\
      &\le \sum_{t=0}^k 2 \log\big(1+1/\bar{v} \cdot\|\mathbf{a}_t\|^2_{\mathcal{D}_{\Sigma^{t}}}\big)\\
      &\le 2\log\Big(\bar{\alpha}^N \cdot\det({\Theta_{k+1}})/\det(\Theta_0)\Big).
    \end{aligned}
  \end{equation}
  Furthermore, based on the recursion~\eqref{ThetaRecursion} of $\Theta_k$, it follows that
  \begin{equation}
    \begin{aligned}
      \Theta_{k+1} &= A[k+1:1]^{-\top}\Theta_0 A[k+1:1]^{-1} \\
      &\hspace{-5pt}+ \sum_{t=0}^kA[k+1:t+1]^{-\top}H_t V^{-1}H_tA[k+1:t+1]^{-1}.
    \end{aligned}
  \end{equation}
 Thus, one can have that
  \begin{equation}
    \begin{aligned}
      &\det(\Theta_{k+1}) \le \Big(1/N \cdot \text{trace}(\Theta_{k+1})\Big)^N\\
      &= \Bigg(1/N \cdot \sum_{i=1}^N\Big(\mathbf{e}_n^\top A[k+1:1]^{-\top}\Theta_0 A[k+1:1]^{-1} \mathbf{e}_n \\
      & \hspace{5pt}+\hspace{-2pt}\sum_{t=0}^k\mathbf{e}_n^\top A[k\hspace{-2pt}+\hspace{-2pt}1\hspace{-2pt}:\hspace{-2pt}t\hspace{-2pt}+\hspace{-2pt}1]^{-\top}H_t V^{-1}H_tA[k\hspace{-2pt}+\hspace{-2pt}1\hspace{-2pt}:\hspace{-2pt}t\hspace{-2pt}+\hspace{-2pt}1]^{-1} \mathbf{e}_n\Big)\Bigg)^N\\
      &\le\Bigg(1/N \cdot \sum_{i=1}^N\Big((\ubar{\alpha}\ubar{\sigma})^{-1} + \sum_{t=0}^k (\ubar{\alpha} \ubar{v})^{-1} \Big)\Bigg)^N\\
      & = \Big((\ubar{\alpha}\ubar{\sigma})^{-1} + (k+1)\cdot (\ubar{\alpha} \ubar{v})^{-1}\Big)^N.
    \end{aligned}
  \end{equation}
  Note that the last inequality is due to the facts i) $\Sigma_0 \le \ubar{\sigma}\cdot \mathbf{I}$; ii) $A[k:t]^{-\top} A[k:t]^{-1} \le \ubar{\alpha}^{-1}\cdot \mathbf{I}$~(see Assumption~\ref{assump:dynamicsBound}); and iii) $H_t^\top V^{-1}H_t \le \ubar{v}^{-1}\cdot \mathbf{I}$ since the specific form of $H_t$ and Assumption~\ref{assump:measurementNoise}.
  As a consequence, it holds that
  \begin{equation}
    \begin{aligned}
      &\log\Big(\bar{\alpha}^N \cdot\det({\Theta_{k+1}})/\det(\Theta_0)\Big)\\
      & \le N \cdot \log\Big(\det(\Sigma_0)^{1/N} \cdot \bar{\alpha} \big((\ubar{\alpha}\ubar{\sigma})^{-1} + (k+1)\cdot (\ubar{\alpha} \ubar{v})^{-1}\big)\Big).
    \end{aligned}
  \end{equation}
  Together with the inequality~\eqref{minIneq}, the proof of Lemma~\ref{lemma:normBound} is completed.
\end{IEEEproof}

With the help of the above Lemma~\ref{lemma:normBound}, we can now continue our proof for the theorem. {Since Lemma~\ref{lemma:lossUpperBound} has guaranteed that the loss of function $r_k \le \bar{\gamma} = 2\sqrt{I \bar{\alpha}} \cdot\|\bm{\phi}_0\|^2, \forall k >0$ Based on the inequality~\eqref{lossIneq}, it follows that
\begin{equation}\label{lossBound}
  \begin{aligned}
    r_k &\le \min\Big\{\bar{\gamma}, \;2\sqrt{N}\beta_k(\delta)\cdot \|\mathbf{a}_k\|_{\mathcal{D}_{\Sigma_{k}}}\Big\} \\
    & \le\kappa \cdot\min\Big\{1, \;2\sqrt{N}\beta_k(\delta)/\sqrt{\bar{v}}\cdot \|\mathbf{a}_k\|_{\mathcal{D}_{\Sigma_{k}}}\Big\} \\
    & \le \kappa \beta'_k(\delta)\cdot \min\Big\{1, \;1/\sqrt{\bar{v}}\cdot \|\mathbf{a}_k\|_{\mathcal{D}_{\Sigma_{k}}}\Big\}.
  \end{aligned}
\end{equation}
In the last two inequalities, we let $\kappa = \max\{\bar{\gamma},\, \sqrt{\bar{v}}\}$ and $\beta'_k(\delta) = \max\{1,\,2\sqrt{2} \beta_k(\delta)\}$. According to the definition~\eqref{betaK} of the non-decreasing sequence $\{\beta_k(\delta)\}_{k \in \mathbb{N}_+}$, it can be seen that the sequence $\{\beta'_k(\delta)\}_{k \in \mathbb{N}_+}$ is also non-decreasing, i.e., $\beta'_k(\delta) \le \beta'_{k+1}(\delta)$. Then, one can have
\begin{equation}\label{finalIneq}
  \begin{aligned}
    &\sum_{k=0}^{K-1} r_k \le  \sqrt{K\cdot \sum_{k=0}^{K-1} r_k^2}\\
     &\hspace{-5pt}\overset{(8.a)}{\le} \kappa \beta'_K(\delta)\cdot \sqrt{K\cdot\sum_{k=0}^{K-1} \min\Big\{1, \;1/\bar{v}\cdot \|\mathbf{a}_k\|^2_{\mathcal{D}_{\Sigma_{k}}}\Big\}}\\
     &\hspace{-5pt}\overset{(8.b)}{\le}\kappa \beta'_K(\delta)\cdot \sqrt{ 2KN} \\
     &\hspace{10pt}\boldsymbol{\cdot}\sqrt{\log\Big(\det(\Sigma_0)^{1/N} \cdot \bar{\alpha}\big((\ubar{\alpha}\ubar{\sigma})^{-1} + K\cdot(\ubar{\alpha} \ubar{v})^{-1}\big)\Big)},
  \end{aligned}
\end{equation}
where $(8.a)$ follows from the inequality~\eqref{lossBound} and $(8.b)$ is due to Lemma~\ref{lemma:normBound}. {Given that \mbox{$\beta'_K(\delta) = \max\{1,\,2\sqrt{2}\beta_K(\delta)\}$} and $\beta_K(\delta) = \mathcal{O}\big(\sqrt{\log(K) }\big)$ in Proposition~\ref{lemma:DUCB}, it can be obtained either $\beta'_K(\delta) = 1$ or $\beta'_K(\delta) = \mathcal{O}\big(\sqrt{\log (K)}\big)$. Therefore, together with the inequality~\eqref{finalIneq}, the statement in Theorem~\ref{theorem1} is proved, i.e., $\sum_{k = 0}^K r_k \le \mathcal{O}\big(\sqrt{K} \log{(K)}\big)$. }
\bibliographystyle{unsrt}
\bibliography{main}

\begin{thebibliography}{10}

\bibitem{tang2005motion}
Zhijun Tang and Umit Ozguner.
\newblock Motion planning for multitarget surveillance with mobile sensor
  agents.
\newblock {\em IEEE Transactions on Robotics}, 21(5):898--908, 2005.

\bibitem{ghaffarkhah2012path}
Alireza Ghaffarkhah and Yasamin Mostofi.
\newblock Path planning for networked robotic surveillance.
\newblock {\em IEEE Transactions on Signal Processing}, 60(7):3560--3575, 2012.

\bibitem{lu2012spoc}
Rongxing Lu, Xiaodong Lin, and Xuemin Shen.
\newblock {SPOC}: A secure and privacy-preserving opportunistic computing
  framework for mobile-healthcare emergency.
\newblock {\em IEEE Transactions on Parallel and Distributed Systems},
  24(3):614--624, 2012.

\bibitem{lu2016cooperative}
Qiang Lu, Qing-Long Han, Botao Zhang, Dongliang Liu, and Shirong Liu.
\newblock Cooperative control of mobile sensor networks for environmental
  monitoring: an event-triggered finite-time control scheme.
\newblock {\em IEEE transactions on cybernetics}, 47(12):4134--4147, 2016.

\bibitem{qian2020real}
Kun Qian and Christian~G. Claudel.
\newblock Real-time mobile sensor management framework for city-scale
  environmental monitoring.
\newblock {\em Journal of Computational Science}, 45, 2020.

\bibitem{mascarich2018radiation}
Frank Mascarich, Taylor Wilson, Christos Papachristos, and Kostas Alexis.
\newblock Radiation source localization in {GPS}-denied environments using
  aerial robots.
\newblock In {\em 2018 IEEE International Conference on Robotics and
  Automation}, pages 6537--6544. IEEE, 2018.

\bibitem{sugiyama2013real}
Hisayoshi Sugiyama, Tetsuo Tsujioka, and Masashi Murata.
\newblock Real-time exploration of a multi-robot rescue system in disaster
  areas.
\newblock {\em Advanced Robotics}, 27(17):1313--1323, 2013.

\bibitem{arnold2018search}
Ross~D Arnold, Hiroyuki Yamaguchi, and Toshiyuki Tanaka.
\newblock Search and rescue with autonomous flying robots through
  behavior-based cooperative intelligence.
\newblock {\em Journal of International Humanitarian Action}, 3(1):1--18, 2018.

\bibitem{abdelkader2013uav}
Mohamed Abdelkader, Mohammad Shaqura, Christian~G Claudel, and Wail Gueaieb.
\newblock A {UAV} based system for real time flash flood monitoring in desert
  environments using lagrangian microsensors.
\newblock In {\em 2013 International Conference on Unmanned Aircraft Systems
  (ICUAS)}, pages 25--34. IEEE, 2013.

\bibitem{rolf2020successive}
Esther Rolf, David Fridovich-Keil, Max Simchowitz, Benjamin Recht, and Claire
  Tomlin.
\newblock A successive-elimination approach to adaptive robotic source seeking.
\newblock {\em IEEE Transactions on Robotics}, 2020.

\bibitem{li2014cooperative}
Shuai Li, Ruofan Kong, and Yi~Guo.
\newblock Cooperative distributed source seeking by multiple robots: Algorithms
  and experiments.
\newblock {\em IEEE/ASME Transactions on Mechatronics}, 19(6):1810--1820, 2014.

\bibitem{fabbiano2014source}
Ruggero Fabbiano, Carlos~Canudas De~Wit, and Federica Garin.
\newblock Source localization by gradient estimation based on {Poisson}
  integral.
\newblock {\em Automatica}, 50(6):1715--1724, 2014.

\bibitem{brinon2015distributed}
Lara Bri{\~n}{\'o}n-Arranz, Luca Schenato, and Alexandre Seuret.
\newblock Distributed source seeking via a circular formation of agents under
  communication constraints.
\newblock {\em IEEE Transactions on Control of Network Systems}, 3(2):104--115,
  2015.

\bibitem{fabbiano2016distributed}
Ruggero Fabbiano, Federica Garin, and Carlos Canudas-de Wit.
\newblock Distributed source seeking without global position information.
\newblock {\em IEEE Transactions on Control of Network Systems}, 5(1):228--238,
  2016.

\bibitem{atanasov2012stochastic}
Nikolay Atanasov, Jerome Le~Ny, Nathan Michael, and George~J Pappas.
\newblock Stochastic source seeking in complex environments.
\newblock In {\em 2012 IEEE International Conference on Robotics and
  Automation}, pages 3013--3018. IEEE, 2012.

\bibitem{azuma2012stochastic}
Shun-ichi Azuma, Mahmut~Selman Sakar, and George~J Pappas.
\newblock Stochastic source seeking by mobile robots.
\newblock {\em IEEE Transactions on Automatic Control}, 57(9):2308--2321, 2012.

\bibitem{atanasov2015distributed}
Nikolay~A Atanasov, Jerome Le~Ny, and George~J Pappas.
\newblock Distributed algorithms for stochastic source seeking with mobile
  robot networks.
\newblock {\em Journal of Dynamic Systems, Measurement, and Control}, 137(3),
  2015.

\bibitem{marchant2012bayesian}
Roman Marchant and Fabio Ramos.
\newblock Bayesian optimisation for intelligent environmental monitoring.
\newblock In {\em 2012 IEEE/RSJ International Conference on Intelligent Robots
  and Systems}, pages 2242--2249. IEEE, 2012.

\bibitem{bai2016information}
Shi Bai, Jinkun Wang, Fanfei Chen, and Brendan Englot.
\newblock Information-theoretic exploration with {Bayesian} optimization.
\newblock In {\em 2016 IEEE/RSJ International Conference on Intelligent Robots
  and Systems}, pages 1816--1822. IEEE, 2016.

\bibitem{miller2015ergodic}
Lauren~M Miller, Yonatan Silverman, Malcolm~A MacIver, and Todd~D Murphey.
\newblock Ergodic exploration of distributed information.
\newblock {\em IEEE Transactions on Robotics}, 32(1):36--52, 2015.

\bibitem{luo2018adaptive}
Wenhao Luo and Katia Sycara.
\newblock Adaptive sampling and online learning in multi-robot sensor coverage
  with mixture of {Gaussian} processes.
\newblock In {\em 2018 IEEE International Conference on Robotics and
  Automation}, pages 6359--6364. IEEE, 2018.

\bibitem{luo2019distributed}
Wenhao Luo, Changjoo Nam, George Kantor, and Katia Sycara.
\newblock Distributed environmental modeling and adaptive sampling for
  multi-robot sensor coverage.
\newblock In {\em 18th International Conference on Autonomous Agents and
  Multi-Agent Systems}, pages 1488--1496, 2019.

\bibitem{benevento2020multi}
Alessia Benevento, Mar{\'\i}a Santos, Giuseppe Notarstefano, Kamran Paynabar,
  Matthieu Bloch, and Magnus Egerstedt.
\newblock Multi-robot coordination for estimation and coverage of unknown
  spatial fields.
\newblock In {\em 2020 IEEE International Conference on Robotics and
  Automation}, pages 7740--7746. IEEE, 2020.

\bibitem{o1978curve}
Anthony O'Hagan.
\newblock Curve fitting and optimal design for prediction.
\newblock {\em Journal of the Royal Statistical Society: Series B
  (Methodological)}, 40(1):1--24, 1978.

\bibitem{2021ICRA}
Bin Du, Kun Qian, Hassan Iqbal, Chris Claudel, and Dengfeng Sun.
\newblock Multi-robot dynamical source seeking in unknown environments.
\newblock In {\em 2021 IEEE International Conference on Robotics and Automation
  (to appear)}. IEEE, 2021.

\bibitem{ould2009distributed}
ElMoustapha Ould-Ahmed-Vall, Douglas~M Blough, Bonnie~Heck Ferri, and George~F
  Riley.
\newblock Distributed global {ID} assignment for wireless sensor networks.
\newblock {\em Ad Hoc Networks}, 7(6):1194--1216, 2009.

\bibitem{du2020jacobi}
Bin Du, Kun Qian, Christian Claudel, and Dengfeng Sun.
\newblock Jacobi-style iteration for distributed submodular maximization.
\newblock {\em arXiv preprint arXiv:2010.14082}, 2020.

\bibitem{li2019boundedness}
Wangyan Li, Zidong Wang, Daniel~WC Ho, and Guoliang Wei.
\newblock On boundedness of error covariances for {Kalman} consensus filtering
  problems.
\newblock {\em IEEE Transactions on Automatic Control}, 65(6):2654--2661, 2019.

\bibitem{battistelli2014kullback}
Giorgio Battistelli and Luigi Chisci.
\newblock {Kullback}--{Leibler} average, consensus on probability densities,
  and distributed state estimation with guaranteed stability.
\newblock {\em Automatica}, 50(3):707--718, 2014.

\bibitem{battistelli2014consensus}
Giorgio Battistelli, Luigi Chisci, Giovanni Mugnai, Alfonso Farina, and Antonio
  Graziano.
\newblock Consensus-based linear and nonlinear filtering.
\newblock {\em IEEE Transactions on Automatic Control}, 60(5):1410--1415, 2014.

\bibitem{cattivelli2010diffusion}
Federico~S Cattivelli and Ali~H Sayed.
\newblock Diffusion strategies for distributed kalman filtering and smoothing.
\newblock {\em IEEE Transactions on Automatic Control}, 55(9):2069--2084, 2010.

\bibitem{habibi2016gradient}
Jalal Habibi, Hamid Mahboubi, and Amir~G Aghdam.
\newblock A gradient-based coverage optimization strategy for mobile sensor
  networks.
\newblock {\em IEEE Transactions on Control of Network Systems}, 4(3):477--488,
  2016.

\bibitem{olfati2005distributed}
R.~Olfati-Saber.
\newblock Distributed kalman filter with embedded consensus filters.
\newblock In {\em 44th IEEE Conference on Decision and Control}, pages
  8179--8184. IEEE, 2005.

\bibitem{olfati2005consensus}
R.~Olfati-Saber and J.~Shamma.
\newblock Consensus filters for sensor networks and distributed sensor fusion.
\newblock In {\em 44th IEEE Conference on Decision and Control}, pages
  6698--6703. IEEE, 2005.

\bibitem{olfati2007distributed}
R.~Olfati-Saber.
\newblock Distributed {Kalman} filtering for sensor networks.
\newblock In {\em 46th IEEE Conference on Decision and Control}, pages
  5492--5498. IEEE, 2007.

\bibitem{atanasov2014joint}
Nikolay Atanasov, Roberto Tron, Victor~M Preciado, and George~J Pappas.
\newblock Joint estimation and localization in sensor networks.
\newblock In {\em 53rd IEEE Conference on Decision and Control}, pages
  6875--6882. IEEE, 2014.

\bibitem{bennetts2013towards}
Victor Manuel~Hernandez Bennetts, Achim~J Lilienthal, Ali~Abdul Khaliq,
  Victor~Pomareda Sese, and Marco Trincavelli.
\newblock Towards real-world gas distribution mapping and leak localization
  using a mobile robot with 3d and remote gas sensing capabilities.
\newblock In {\em 2013 IEEE International Conference on Robotics and
  Automation}, pages 2335--2340. IEEE, 2013.

\bibitem{jiang2019source}
Xiangyuan Jiang, Shuai Li, Bing Luo, and Qinghao Meng.
\newblock Source exploration for an under-actuated system: A control-theoretic
  paradigm.
\newblock {\em IEEE Transactions on Control Systems Technology},
  28(3):1100--1107, 2019.

\bibitem{li2014multi}
Shuai Li, Yi~Guo, and Brian Bingham.
\newblock Multi-robot cooperative control for monitoring and tracking dynamic
  plumes.
\newblock In {\em 2014 IEEE International Conference on Robotics and
  Automation}, pages 67--73. IEEE, 2014.

\bibitem{jiang2018plume}
Xiangyuan Jiang and Shuai Li.
\newblock Plume front tracking in unknown environments by estimation and
  control.
\newblock {\em IEEE Transactions on Industrial Informatics}, 15(2):911--921,
  2018.

\bibitem{morton2019revival}
Keith~W Morton.
\newblock {\em Revival: Numerical Solution Of Convection-Diffusion Problems
  (1996)}.
\newblock CRC Press, 2019.

\bibitem{abbasi2011improved}
Yasin Abbasi-Yadkori, D{\'a}vid P{\'a}l, and Csaba Szepesv{\'a}ri.
\newblock Improved algorithms for linear stochastic bandits.
\newblock {\em Advances in Neural Information Processing Systems},
  24:2312--2320, 2011.

\end{thebibliography}

\end{document}